\documentclass[journal]{IEEEtran}
\PassOptionsToPackage{numbers, compress}{natbib}

\ifCLASSINFOpdf
\else
\fi
\usepackage{times}
\usepackage{epsfig}
\usepackage{graphicx}
\usepackage{amsmath}
\usepackage{amssymb}
\usepackage{subfigure}
\usepackage{color}
\usepackage{cite}
\usepackage{overpic}
\usepackage{algorithm}
\usepackage{algorithmic}
\usepackage{graphics}
\usepackage{enumitem} 

\usepackage{xspace}

\makeatletter
\DeclareRobustCommand\onedot{\futurelet\@let@token\@onedot}
\def\@onedot{\ifx\@let@token.\else.\null\fi\xspace}

\def\eg{\emph{e.g}\onedot} 
\def\ie{\emph{i.e}\onedot} 
\def\cf{\emph{c.f}\onedot}

\def\resp{\emph{resp}\onedot}
\makeatother

\newcommand{\HL}[1]{\textcolor[rgb]{1.00,0.00,0.00}{#1}}
\newcommand{\JP}[1]{\textcolor[rgb]{0.00,0.00,1.00}{#1}}
\newcommand{\JLP}[1]{\textcolor[rgb]{0.00,1.00,0.00}{#1}}

\begin{document}
%
\title{Super Diffusion for Salient Object Detection}
%
%
%

\author{Peng~Jiang,
        Zhiyi~Pan,
        Nuno~Vasconcelos,~\IEEEmembership{Fellow,~IEEE,}
        Baoquan~Chen,~\IEEEmembership{Senior Member,~IEEE,}
        and~Jingliang~Peng,~\IEEEmembership{Member,~IEEE}
\thanks{This work was partially funded by the National Natural Science Foundation of China (NSFC Grants No.s 61702301, 61472223 and 61872398) and the China Postdoctoral Science Foundation (Grant No. 2017M612272).}
\thanks{P. Jiang, Z. Pan and J. Peng are with Shandong University, China. (E-mails:\{sdujump, jingliap, panzhiyi1996\}@gmail.com). N. Vasconcelos is with University of California, San Diego, USA. (E-mail: nvasconcelos@ucsd.edu). B. Chen is with Peking University, China. (E-mail: baoquan.chen@gmail.com)}
\thanks{J. Peng is the corresponding author.}}
%
%

\markboth{Journal of \LaTeX\ Class Files,~Vol.~X, No.~X, November~2018}%
{JIANG \MakeLowercase{\textit{et al.}}: Super Diffusion for Salient Object Detection}
%



\maketitle

\begin{abstract}
One major branch of saliency object detection methods is diffusion-based which construct a graph model on a given image and diffuse seed saliency values to the whole graph by a diffusion matrix.  While their performance is sensitive to specific feature spaces and scales used for the diffusion matrix definition, little work has been published to systematically promote the robustness and accuracy of salient object detection under the generic mechanism of diffusion.

In this work, we firstly present a novel view of the working mechanism of the diffusion process based on mathematical analysis, which reveals that the diffusion process is actually computing the similarity of nodes with respect to the seeds based on diffusion maps. Following this analysis, we propose super diffusion, a novel inclusive learning-based framework for salient object detection, which makes the optimum and robust performance by integrating a large pool of feature spaces, scales and even features originally computed for non-diffusion-based salient object detection. A closed-form solution of the optimal parameters for the integration is determined through supervised learning.

At the local level, we propose to promote each individual diffusion before the integration. Our mathematical analysis reveals the close relationship between saliency diffusion and spectral clustering. Based on this, we propose to re-synthesize each individual diffusion matrix from the most discriminative eigenvectors and the constant eigenvector (for saliency normalization).

The proposed framework is implemented and experimented on prevalently used benchmark datasets, consistently leading to state-of-the-art performance.
\end{abstract}


%
\IEEEpeerreviewmaketitle

\section{Introduction}
%
%
%
%
\IEEEPARstart{T}he aim of saliency detection is to identify the most salient pixels or regions in a digital image which attract humans' first visual attention. Results of saliency detection can be applied to other computer vision tasks such as image resizing, thumbnailing, image segmentation and object detection. Due to its importance, saliency detection has received intensive research attention resulting in many recently proposed algorithms.

In the field of saliency detection, two branches have developed, which are visual saliency detection~\cite{v1,v4,v5,v7,v8,v10,v12,v13,v14,v15,v16,v17} and salient object detection~\cite{s1,s2,s3,s4,s5,s7,s9,s10,s11,s12,s14,s15,s16,s17,s18,s19,s20,s21,s22,s23,s24,s28,grab,EQCUTS,GP,SP,RW}. While the former tries to predict where the human eyes focus on, the latter aims to detect the whole salient object in an image. Saliency in both branches can be computed in a bottom-up fashion using low-level features
~\cite{v1,v4,v7,v8,v13,v14,v15,v16,v17,s1,s2,s5,s7,s11,s12,s14,s19,s21,s22,s23,s24,s28,GP,grab,EQCUTS,SP,RW}, in a top-down fashion by training with certain samples driven by specific tasks~\cite{s16,v12,s4,v5,s18,s17,v10,s20}, or in a way of combining both low-level and high-level features~\cite{s3,s9,s10,s15}. In this work, we focus on salient object detection and utilize both high-level training and low-level features.

Salient object detection algorithms usually generate bounding boxes, binary foreground and background segmentation, or saliency maps which indicate the saliency likelihood of each pixel. Over the past several years, contrast-based methods~\cite{s1,s2,s7,s9} significantly promote the benchmark of salient object detection. However, these methods usually miss small local salient regions or bring some outliers such that the resultant saliency maps tend to be nonuniform. To tackle these problems, diffusion-based methods~\cite{s11,s12,s18,s22,grab,EQCUTS,GP,SP,RW} use diffusion matrices to propagate saliency information of seeds to the whole salient object.
While most of them focus on constructing good graph structures, generating good seed vectors and/or controlling the diffusion process, they have not yet made sufficient effort in analyzing the fundamental working mechanism of the diffusion process and accordingly addressing the inherent problems with the diffusion-based approaches.

The existent diffusion-based methods more or less follow a restricted framework, \ie, a specific diffusion matrix is defined in specific feature space and scale based on a specific graph structure, usually with the seed saliency vector computed according to specific color-space heuristics. As a result, they usually lack in extensibility and robustness.
This has motivated our search in this work for an inclusive and extensible diffusion-based framework that incorporates a large pool of feature spaces, scales, and seeds for robust performance. Major contributions of this work reside in the following aspects.

\begin{itemize}

\item \textbf{Novel interpretation of the diffusion mechanism}. Through eigen-analysis of the diffusion matrix, we find that: 1) the saliency of a node (called focus node) is equal to a weighted sum of all the seed saliency values, with the weights determined by the similarity in diffusion map between the focus node and each seed node, and 2) since the diffusion map is formed by the eigenvectors and eigenvalues of the diffusion matrix, the process of saliency diffusion has a close relationship with spectral clustering. This novel interpretation provides the foundation for the novel framework and methods proposed in this work.

\item \textbf{Super diffusion framework for salient object detection}.  We propose an inclusive and extensible framework, named super diffusion, for salient object detection, which computes the optimal diffusion matrix by exploiting a pool of feature spaces, scales and even saliency features originally computed for non-diffusion-based salient object detection. The parameters for the optimal integration are derived in a closed-form solution through supervised learning. This contrasts with traditional diffusion-based methods that define the diffusion matrices and seeds with high specificity, compromising the robustness of performance.

\item \textbf{Local refinement of saliency diffusion}.  We propose to promote each individual saliency diffusion scheme prior to its integration into the overall super diffusion framework. Based on the close relationship between saliency diffusion and spectral clustering, the promotion is achieved by re-synthesizing an individual diffusion matrix from the most discriminative eigenvectors and the constant eigenvector (for saliency normalization). In addition, we propose efficient and effective ways to compute seed vectors based on background and foreground priors.

\end{itemize}

It should be noted that an initial version of this work was published as a conference paper~\cite{GP}, which has been extended to this journal version mainly in the following aspects: 1) proposal of the super diffusion framework with full-length explanation, 2) significantly extended experiments to evaluate the proposed framework in concrete implementation, and 3) more comprehensive coverage and analysis of related works.

\section{Related Works}

Diffusion-based salient object detection methods(\eg, ~\cite{s11,s12,s18,grab,GP,SP,RW,EQCUTS}) share the same main formula:
 \begin{equation}
 \begin{aligned}
    \mathbf{y} = \mathbf{A}^{-1}\mathbf{s},
 \end{aligned}
 \label{eq:eq1}
 \end{equation}
where $\mathbf{A}^{-1}$ is the diffusion matrix (also called ranking matrix or propagation matrix), $\mathbf{s}$ is the seed vector (diffusion seed),
and $\mathbf{y}$ is the final saliency vector to be computed.
Here $\mathbf{s}$ usually contains preliminary saliency information of a portion of nodes, that is to say, usually $\mathbf{s}$ is not complete and we need to propagate the partial saliency information in $\mathbf{s}$ to the whole salient region based on a graph structure to obtain the final saliency map, $\mathbf{y}$~\cite{s18}. The diffusion matrix $\mathbf{A}^{-1}$ is designed to fulfill this task. The existent methods mostly focus on how to construct the graph structure, how to generate the seed and/or how to control the diffusion process. Accordingly, we review them based on their approaches to the three sub-problems, respectively, in the following sub-sections.

\subsection{Graph Construction}

A diffusion-based salient object detection algorithm needs to firstly construct a graph structure on a given image for the definition of diffusion matrix. Specifically, it segments the given image into $N$ superpixels first by an algorithm such as SLIC~\cite{a3} or ERS~\cite{ERS}, and then constructs a graph $G = (V,E)$ with superpixels as nodes $v_i$, $1 \leq i \leq N$, and undirected links between node pairs $(v_i, v_j)$ as edges $e_{ij}$, $1 \leq i, j \leq N$, to define the adjacency. Note that superpixels but not pixels are usually used as nodes for efficiency and stability considerations.

Straightforwardly, two nodes are connected by an edge in the graph if they are contiguous in the image. In order to capture relationship between nodes farther on the image, some works~\cite{s11,s12,s18,grab,GP,RW} connect a node to not only its directly contiguous neighbors, but also its 2-hop and even up to 5-hop neighbors~\cite{EQCUTS}. Besides, some works~\cite{s11,s12,s18,grab,GP,SP,RW} make a close-loop graph by connecting the nodes at the four borders of the image to each other. As a result, the distance between two nodes close to two different borders will be shortened by a path through borders. Work~\cite{grab} connects each node to all the nodes at the four borders to increase the connectivity of the graph, which provides certain robustness to noise.

The weight $w_{ij}$ of the edge $e_{ij}$ which encodes the similarity between linked nodes usually is defined as
\begin{equation}
\begin{aligned}
w_{ij} =  e^{-\frac{\|v_{i}-v_{j}\|_2}{\sigma^2}}
\end{aligned}
\label{eq:resp}
\end{equation}
where $v_{i}$ and $v_{j}$ represent the mean feature value of two nodes respectively, and $\sigma$ is a scale parameter that controls the strength of the weight. All the mentioned diffusion-based methods use the CIE LAB color space as feature space. Finally, the affinity matrix is defined as $\mathbf{W} = [w_{ij} ]_{N{\times}N}$ with $w_{ij}$ computed by Eq.~\ref{eq:resp} if $i=j$ or edge $e_{ij}$ exists in the graph and assigned 0 otherwise; the degree matrix is defined as $\mathbf{D} = diag\{d_{11},...,d_{NN}\}$, where $d_{ii} = \sum_j{w_{ij}}$.

\subsection{Diffusion Matrix and Seed Computation}

Different algorithms derive diffusion matrices and seed vectors in different ways. Works~\cite{s11,grab,RW} use inverse Laplacian matrix $\mathbf{L}^{-1}$ as the diffusion matrix. Correspondingly, the formula of saliency diffusion is
 \begin{equation}
 \begin{aligned}
    \mathbf{y} = &\mathbf{L}^{-1}\mathbf{s}
 \end{aligned}
 \label{eq:eq2}
 \end{equation}
 where $\mathbf{L}=\mathbf{D}-\mathbf{W}$.
Works~\cite{s18,SP} use inverse normalized Laplacian matrix $\mathbf{L_{rw}}^{-1}$ as the diffusion matrix which normalizes weights by degrees of nodes when computing similarity. Correspondingly, the formula of saliency diffusion is
 \begin{equation}
 \begin{aligned}
    \mathbf{y} = &\mathbf{L_{rw}}^{-1}\mathbf{s}
 \end{aligned}
 \label{eq:eq3}
 \end{equation}
where $\mathbf{L_{rw}}=(\mathbf{I}-\mathbf{D}^{-1}\mathbf{W})=\mathbf{D}^{-1}(\mathbf{D}-\mathbf{W})$.

While works~\cite{s11,grab,RW,SP} use binary background and foreground indication vectors as the seed vectors in two stages, respectively, work~\cite{s18} computes $\mathbf{s}$ by combining hundreds of saliency features $\mathbf{F}$ with learned weight $\mathbf{w}$ ($\mathbf{s}=\mathbf{F}\mathbf{w}$).

Different from the above-mentioned methods, work~\cite{s12} duplicates the superpixels around the image borders as the virtual background absorbing nodes, and sets the inner nodes as transient nodes. Then, the entry of seed vector $s_i=1$ if node $v_i$ is a transient node and $s_i=0$ otherwise. Correspondingly, the formula of saliency diffusion is
 \begin{equation}
 \begin{aligned}
	 \mathbf{y} = &(\mathbf{I}-\mathbf{P})^{-1}\mathbf{s}=\mathbf{L_{rw}}^{-1}\mathbf{s}
 \end{aligned}
 \label{eq:eq4}
 \end{equation}
where $\mathbf{P}=\mathbf{D}^{-1}\mathbf{W}$ and $\mathbf{P}$ is called transition matrix. Note that Eq.~\ref{eq:eq4} is derived from but not identical to the original formula in reference~\cite{s12} and the derivation process is described in Appendix~\ref{sect:appendix}.

In general, the existing diffusion-based salient object detection methods derive their diffusion matrices from the basic form of Laplacian matrix. As a result, their performance is restricted by the Laplacian matrix that makes the performance sensitive to the scale parameter and the feature space used for the matrix construction.

\subsection{Diffusion Process Control}

Applying Eq.~\ref{eq:eq1} for once to complete the salient object detection task may not produce satisfactory results, as the seed saliency information may diffuse to the non-salient region or may not diffuse to the whole salient region. One common way to control the diffusion process is applying multi-stage diffusion instead of one-stage diffusion. Works~\cite{s11,grab,SP,RW} diffuse to estimate a non-saliency map using the background prior, and reverses and thresholds the map to get the most salient seed nodes at the first stage; they conduct another pass of diffusion at the second stage with the seed saliency estimated at the first stage. Work~\cite{SP} further divides each pass of diffusion into a sequence of steps that, instead of computing saliency of all nodes at once, estimates saliency of a subset of nodes as selected according to certain rules.  Though effective to a certain extent, these approaches lack in theoretical support and may not be robust in general.

To summarize, researchers have devised good ways to construct the graph structures, the diffusion matrices and the seed vectors exploiting effective heuristics and priors. In this work, we step further to explore novel views of the fundamental diffusion mechanism and, accordingly, make systematic promotion of the diffusion-based salient object detection performance.

\section{Diffusion Re-Interpreted} \label{sect:diffusionmap}

As discussed before, a diffusion-based salient objection detection algorithm (\eg, \cite{s11,s12,s18,grab,SP,RW,EQCUTS}) usually defines the diffusion matrix by a certain form of the Laplacian matrix, denoted by $\mathbf{A}$, which usually is positive semi-definite. Thus, $\mathbf{A}$ can be decomposed as $\mathbf{A}={\mathbf{U}}{\mathbf{\Lambda}}{\mathbf{U}}^{T}$ where $\mathbf{\Lambda}$ is a diagonal matrix formed from the eigenvalues $\lambda_{l}$, $l=1,2,\ldots,N$, and the columns of $\mathbf{U}$ are the corresponding eigenvectors $\mathbf{u}_{l}$, $l=1,2,\ldots,N$. According to spectral decomposition theories, each element, $\tilde a(i,j)$, of $\mathbf{A}^{-1}$ can then be expressed as
 \begin{equation}
 \begin{aligned}
     \tilde a(i,j)=\sum_{l=1}^N \lambda_{l}^{-1} \mathbf{u}_{l}(i)\mathbf{u}_{l}(j).
 \end{aligned}
 \end{equation}
and each entry, $y_i$, of $\mathbf{y}$ as
 \begin{equation}
 \begin{aligned}
    \mathbf{y}_i &= \sum_{j=1}^N \mathbf{s}_j \sum_{l=1}^N \lambda_{l}^{-1} \mathbf{u}_{l}(i)\mathbf{u}_{l}(j) \\
    &= \sum_{j=1}^N \mathbf{s}_j\left<\mathbf{\Psi}_{i}, \mathbf{\Psi}_{j}\right>,
 \end{aligned}
 \label{eq:eq6}
 \end{equation}
 \begin{equation}
 \begin{aligned}
 	\mathbf{\Psi}_{i} = [\lambda_{1}^{-\frac{1}{2}}\mathbf{u}_{1}(i),...,\lambda_{N}^{-\frac{1}{2}}\mathbf{u}_{N}(i)]
 \end{aligned}
 \label{eq:eq7}
 \end{equation}
where $\left<\cdot,\cdot\right>$ is the inner product operation.
According to the reference~\cite{a4}, $\mathbf{\Psi}_{i}$ is called diffusion map (diffusion map at time $t=-\frac{1}{2}$ to be more exactly) at the $i$-th data point (node).

Based on Eq.s~\ref{eq:eq6} and \ref{eq:eq7}, we make a novel interpretation of the working mechanism of diffusion-based salient object detection: the saliency of a node (called focus node) is determined by all the seed saliency values in the form of weighted sum, with each weight determined by diffusion map similarity (measured by inner product) between the corresponding seed node and the focus node. In other words, seed nodes having more diffusion map similarity to the focus node will influence more on the focus node's saliency. In a nutshell, diffusion maps are key functional elements for the diffusion.

According to Eq.s~\ref{eq:eq6} and \ref{eq:eq7}, nodes with similar (\resp, distinct) diffusion maps tend to obtain similar (\resp, distinct) saliency values. Therefore, the process of saliency diffusion is closely related to the clustering of the nodes based on their diffusion maps. Further, diffusion maps are derived from the eigenvalues and eigenvectors of the diffusion matrix, \ie, we form a matrix by putting the weighted eigenvectors in columns and each row of the matrix gives one node's diffusion map (see Eq.~\ref{eq:eq7}). As such, the diffusion-map-based clustering is almost identical in form to the standard spectral clustering of the nodes~\cite{a1,a2}.

According to~\cite{a1,a2,a9}, the spectral clustering performance tends to be sensitive to the scale parameter $\sigma$ and the feature space used for computing the Laplacian matrix (see Eq.~\ref{eq:resp}), and only a subset of the eigenvectors are the most discriminative while the rest are less discriminative or even cause confusions to the clustering. Due to the close relationship between spectral clustering and saliency diffusion, we foresee that the limitations of the spectral clustering also limit the performance of the saliency diffusion. As such, we address these limitations in this work to fundamentally promote the performance of saliency diffusion.

\section{Super Diffusion}
\subsection{Generic Framework}\label{sect:integration}

We propose a framework that systematically integrates diffusion maps originally derived from various diffusion matrices and seed vectors originally derived by various heuristics, so as to get rid of the sensitiveness of traditional diffusion-based salient object detection methods to specific feature spaces, scales and heuristics. We call this framework \emph{super diffusion}.

For the $i$-th, $1 \leq i \leq N$, node, we may integrate various diffusion maps by defining

 \begin{equation}
 \begin{aligned}
 	\hat{\mathbf{\Psi}}_{i} = [\alpha_1\mathbf{\Psi}_{i}^1, \alpha_2\mathbf{\Psi}_{i}^2,\cdots,\alpha_{M}\mathbf{\Psi}_{i}^M]
 \end{aligned}
 \label{eq:diffmap}
 \end{equation}
where $\mathbf{\Psi}_{i}^j$, $1 \leq j \leq M$, is a diffusion map computed from diffusion matrix $\mathbf{A}_j^{-1}$ at the $i$-th node, and $\alpha_j$ is its weight to be determined. Correspondingly, we may formulate a matrix

 \begin{equation}
 \begin{aligned}
    \mathbf{A}_I^{-1} &= \mathbf{U}_I \mathbf{\Lambda}_I \mathbf{U}_I^T \\
    &= \mathbf{U}_I\left[
	\begin{array}{ccc}
	\alpha_{1}\mathbf{\Lambda}_1^{-1}&\cdots&\cdots\\
	\vdots&\ddots&\vdots\\
	\cdots&\cdots&\alpha_{M}\mathbf{\Lambda}_{M}^{-1}\\
	\end{array}
	\right]\mathbf{U}_I^T,
 \end{aligned}
 \label{eq:lapinte}
 \end{equation}
where $\mathbf{U}_I = \left[\mathbf{U}_1, \cdots, \mathbf{U}_{M} \right]$, $\mathbf{\Lambda}_I$ is a diagonal matrix and $\mathbf{A}_j={\mathbf{U}_j}{\mathbf{\Lambda}_j}{\mathbf{U}_{j}^{T}}$ is the eigen decomposition of $\mathbf{A}_j$.  Applying $\mathbf{A}_I^{-1}$ to a seed vector, $\mathbf{s}$, we obtain the saliency vector $\mathbf{y} = \mathbf{A}_I^{-1}\mathbf{s}$ with its entry, $y_i$, $1 \leq i \leq N$, expressed as $y_i = \sum_{j=1}^N s_j\left<\hat{\mathbf{\Psi}}_{i}, \hat{\mathbf{\Psi}}_{j}\right>$.

Further, we may integrate various seed vectors, $\mathbf{s}^1,\mathbf{s}^2,\cdots,\mathbf{s}^K$, by
 \begin{equation}
 \begin{aligned}
 \mathbf{s}_I &= [\mathbf{s}^1,\mathbf{s}^2,\cdots,\mathbf{s}^K][\beta_1,\beta_2,\cdots,\beta_K]^T
 \end{aligned}
 \label{eq:superseed}
 \end{equation}
with $\beta_k$, $1 \leq k \leq K$, being the weights to be determined.

Applying $\mathbf{A}_I^{-1}$ to $\mathbf{s}_I$, we obtain the saliency vector,
 \begin{equation}
 \begin{aligned}
    \mathbf{y} &= \mathbf{A}_I^{-1}\mathbf{s}_I\\
    &= \sum_{i=1}^{M}\sum_{j=1}^{K} \alpha_i \beta_j \mathbf{U}_i \mathbf{\Lambda}_i^{-1} \mathbf{U}_i^T \mathbf{s}^j \\
    &= \sum_{i=1}^{M}\sum_{j=1}^{K} \alpha_i \beta_j \mathbf{A}_i^{-1} \mathbf{s}^j\\
    &= \sum_{i=1}^{M}\sum_{j=1}^{K} \alpha_i \beta_j \mathbf{y}^{i,j}\\
    &= \mathbf{H} \mathbf{w}^T,
 \end{aligned}
 \label{eq:newdiffu}
 \end{equation}
where $\mathbf{A}_i^{-1}=\mathbf{U}_i\mathbf{\Lambda}_i^{-1}\mathbf{U}_i^T$,
$\mathbf{y}^{i,j}=\mathbf{A}_i^{-1} \mathbf{s}^j$,
$\mathbf{H}=[\mathbf{y}^{1,1}, \cdots, \mathbf{y}^{1,K}, \mathbf{y}^{2,1}, \cdots, \mathbf{y}^{2,K}, \cdots, \mathbf{y}^{M,1}, \cdots, \mathbf{y}^{M,K}]$, and $\mathbf{w}=[\alpha_1 \beta_1, \cdots, \alpha_1 \beta_K, \alpha_2 \beta_1, \cdots, \alpha_2 \beta_K, \cdots, \alpha_M \beta_1, \cdots, \\\alpha_M \beta_K]$. With $\mathbf{A}_i^{-1}$, $1 \leq i \leq M$, and $\mathbf{s}^j$, $1 \leq j \leq K$, given, the variables of this system are $\alpha_i$, $1 \leq i \leq M$, and $\beta_j$, $1 \leq j \leq K$. In other words, the degree of freedom (DOF) for our solution is $M+K$. In order to increase the room for optimization, we increase the DOF to $M \times K$ by replacing $\mathbf{w}$ in Eq.~\ref{eq:newdiffu} with $\mathbf{w}=[w_1, w_2, \cdots, w_{M \times K}]$ and solving for $w_i$, $1 \leq i \leq M \times K$, instead.

We determine the weighting vector, $\mathbf{w}$, by supervised learning from a training set of $L$ samples, with the loss function defined as
 \begin{equation}
 \begin{aligned}
    J &= \sum_{i=1}^L(\mathbf{y}(i)-\mathbf{y_{gt}}(i))^2\\
    &= \sum_{i=1}^L(\mathbf{H}(i)\mathbf{w}^T-\mathbf{y_{gt}}(i))^2,
 \end{aligned}
 \label{eq:loss}
 \end{equation}
where $\mathbf{y}(i)$, $\mathbf{y_{gt}}(i)$ and $\mathbf{H}(i)$ are the computed saliency vector, the ground-truth binary saliency vector and the $\mathbf{H}$ matrix for the $i$-th training sample, respectively. As $J$ is convex, the optimal $\mathbf{w}$ has a closed-form expression of
 \begin{equation}
 \begin{aligned}
    \mathbf{w} &= \frac{\sum_{i=1}^L \mathbf{H}(i)^{T}\mathbf{y_{gt}}(i)}{\sum_{i=1}^L \mathbf{H}(i)^T \mathbf{H}(i)}.
 \end{aligned}
 \label{eq:w}
 \end{equation}

 \subsection{Local Refinement}\label{sect:measures}

While the proposed framework in Sec.~\ref{sect:integration} promotes the robustness by optimally integrating various diffusion matrices and seeds, each individual diffusion matrix on its own may be optimized as well.

As discussed in Sec.~\ref{sect:diffusionmap}, only a subset of $\mathbf{A}$'s eigenvectors are the most discriminative. Thus, in order to increase the discriminative power of the diffusion maps associated with each specific $\mathbf{A}_i$, $1 \leq i \leq M$, in Sec.~\ref{sect:integration},
we are motivated to keep only the most discriminative while discarding the rest of its eigenvectors. Specifically, we refine each individual $\mathbf{A}_i^{-1}$ by re-synthesizing it from $\mathbf{A}_i$'s most discriminative eigenvectors followed by a normalization step, as detailed in the following subsections. We call this process local refinement for short.

In practice, we first refine each individual diffusion matrix, $\mathbf{A}_i^{-1}$, and then use the refined diffusion matrices to compute all the saliency values in matrix $\mathbf{H}$ in Eq.~\ref{eq:newdiffu} and $\mathbf{H}(i)$ in Eq.s~\ref{eq:loss} and~\ref{eq:w}. Regarding the choice of $\mathbf{A}_i$, $1 \leq i \leq M$, we use a slightly modified $\mathbf{L_{rw}}$, $\tilde{\mathbf{L_{rw}}}$(\cf Sec.~\ref{sect:constant}), as the basic form and define a series of diffusion matrices by varying the feature space and scale parameter when computing the edge weights (\cf Eq.~\ref{eq:resp}). Our choice is motivated by the fact that $\mathbf{L_{rw}}$ often leads to better intra-cluster coherency and clustering consistency than $\mathbf{L}$ for spectral clustering (\cf~\cite{a2}).
 
\subsubsection{Constant Eigenvector}\label{sect:constant}

The eigenvalues, $\lambda_l$, and eigenvectors, $\mathbf{u}_l$, $1 \leq l \leq N$, of $\mathbf{L_{rw}}$ (the same for $\mathbf{L}$) are ordered such that $0=\lambda_1 \leq \lambda_2 \leq \ldots \leq \lambda_N$ with $\mathbf{u}_1=\mathbf{1}$~\cite{a1}.
Some works (\eg,~\cite{s11}) avoid zero eigenvalues by approximately setting $\tilde{\mathbf{L_{rw}}}=\mathbf{D}^{-1}(\mathbf{D}-0.99W)$ such that $\tilde{\mathbf{L_{rw}}}$ is always invertible. Assuming $\tilde{\lambda}_l$ and $\tilde{\mathbf{u}_l}$, $1 \leq l \leq N$, are the corresponding eigenvalues and eigenvectors of $\tilde{\mathbf{L}_{rw}}=\mathbf{D}^{-1}(\mathbf{D}-0.99\mathbf{W})$, it can be proven that $\tilde{\mathbf{u}_l}=\mathbf{u}_l$ and $\tilde{\lambda}_l=0.99\lambda_l+0.01$. Thus, $0.01=\tilde{\lambda}_1 \leq \tilde{\lambda}_2 \leq \ldots \leq \tilde{\lambda}_N$ with $\tilde{\mathbf{u}}_1=\mathbf{1}$. 

The constant eigenvector $\tilde{\mathbf{u}}_1$ contains no discriminative information. Thus, we discard it and re-synthesize the diffusion matrix, as done in our early conference version~\cite{GP} of this work. But novelly, we reuse the constant eigenvector later in Sec.~\ref{sect:normalization} for normalization of saliency.

\subsubsection{Eigengap}

\begin{figure}[!t]
 \begin{center}
     \subfigure{
    \put(70,-12){\includegraphics[width=0.1\textwidth, height=0.23\textwidth]{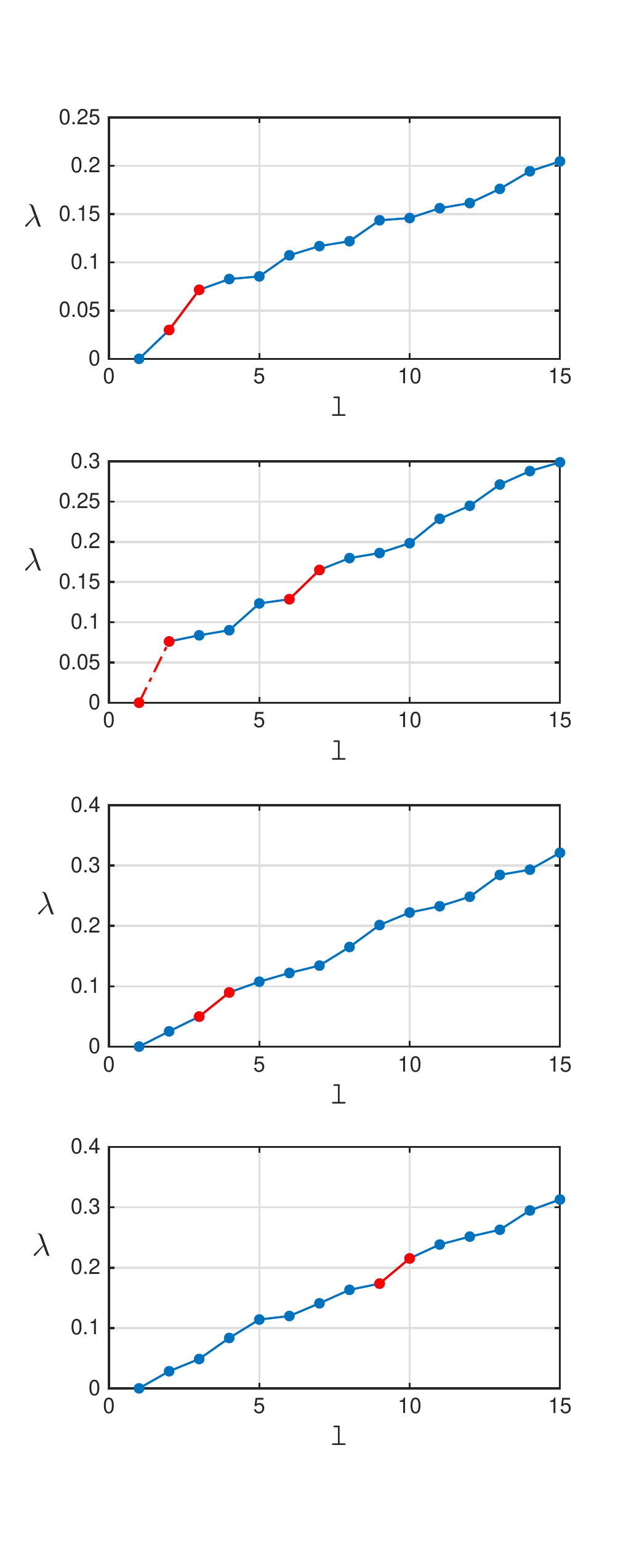}}\put(-118,0)
    {\includegraphics[width=0.38\textwidth]{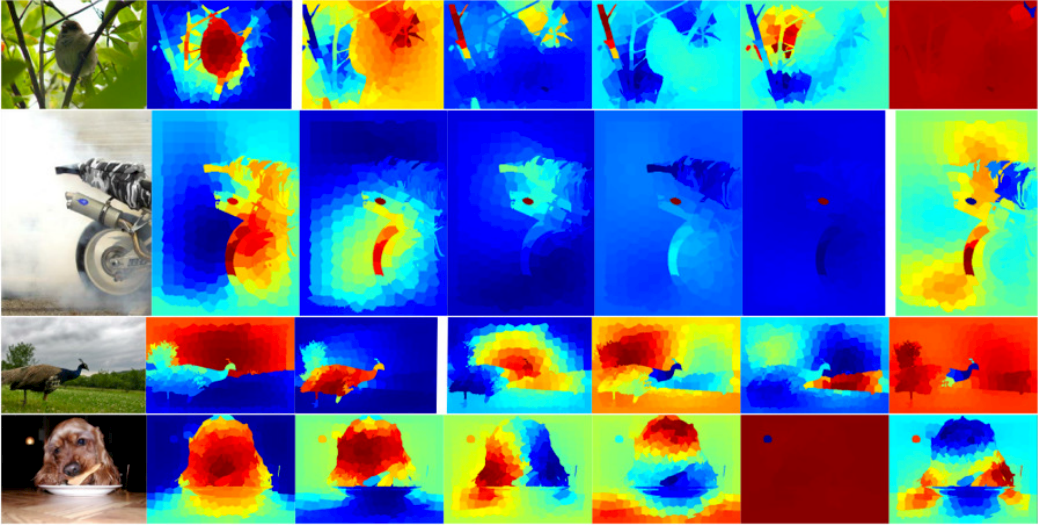}}
     }
 \end{center}
 \vspace{-8mm}
 {\small\hspace{1mm} SRC  \hspace{4mm} $u_{2}$  \hspace{4mm}  $u_{3}$  \hspace{4mm}  $u_{4}$  \hspace{5mm}  $u_{5}$ \hspace{5mm} $u_{6}$ \hspace{5mm}  $u_{7}$    \hspace{8mm} $\lambda$}
 \vspace{2mm}
 \caption{Visualization of normalized eigenvectors by color coding. Pixels in each node are assigned a single color and nodes with similar values in an eigenvector are colored similarly. The eight columns show the source images (SRC), the corresponding eigenvectors ($\mathbf{u}_2$-$\mathbf{u}_7$) and eigenvalue curves ($\lambda$), respectively. We use a white margin between successive eigenvectors to indicate an eigengap (all the eigenvectors, $\mathbf{u}_2$ to $\mathbf{u}_7$, are before the eigengap, if there is no white margin in that row.). Besides, on the eigenvalue curves, we use red solid segments to indicate the final eigengaps and a red dash segment to indicate an initial eigengap of $r=2$ to be reset. Ground-truth saliency of the source images are shown in Fig.~\ref{fig:final}.}
 \label{fig:fig2}
 \end{figure}

In each diffusion matrix, except $\mathbf{u}_1$ that is a constant vector, the more $\mathbf{u}_l$ ($l \in [2,N]$) is to the front of the ordered array, the more indicative it usually is for the clustering. For instance, we visualize in Fig.~\ref{fig:fig2} a leading portion (excluding $\mathbf{u}_1$) of the ordered array of eigenvectors for each of four sample images. From Fig.~\ref{fig:fig2}, we see that, for each sample image, the first few eigenvectors well indicate node clusters while the later ones often convey less or even confusing information about the clustering. The key is how to determine the exact cutting point before which the eigenvectors should be kept and after which discarded.

In practice, $\mathbf{L_{rw}}$ (the same for $\mathbf{L}$) often exhibits an eigengap, \ie, a few of its eigenvalues before the eigengap are much smaller than the rest. Specifically, we denote the eigengap as $r$ and define it as
 \begin{equation}
 \begin{aligned}
	r&=\underset{l}{\operatorname{argmax}}|\Delta{\Upsilon}_l|,\\
	\Delta{\Upsilon}_l& = \lambda_l-\lambda_{l-1}, ~ l=2,\ldots,N.
 \end{aligned}
 \label{eq:eigen}
 \end{equation}
Usually, Eq.~\ref{eq:eigen} is called eigengap heuristic. According to~\cite{a2}, some leading eigenvectors (except $u_1$) before the eigengap are usually good cluster indicators which can capture the data cluster information with good accuracy (as observed in Fig.~\ref{fig:fig2}),
meanwhile the location of the eigengap often indicates the right number of data clusters.
Further, the larger the difference between the two successive eigenvalues at the eigengap is, the more important the leading eigenvectors are, since $u_l$ is weighted by $\lambda_l^{-\frac{1}{2}}$ in diffusion map $\mathbf{\Psi}$ (\cf Eq.~\ref{eq:eq7}).
Ideally, the eigenvalues before the eigengap are close to zero while the rest are much larger, which means that the leading eigenvectors (except $u_1$) will dominate the behavior of the diffusion map.

With the eigengap identified, we then keep only the eigenvectors prior to the eigengap, which are usually the most discriminative ones for the task of node clustering. It may sometimes happen that $r=2$ according to Eq.~\ref{eq:eigen}, meaning that all the eigenvectors will be filtered out. In this case, we assume the position of the second largest $|\Delta{\Upsilon}_l|$ as the eigengap.

\subsubsection{Discriminability}\label{sect:discriminability}

In some cases, an eigenvector may only distinguish a tiny region from the background, \eg, $\mathbf{u}_{5}$ , $\mathbf{u}_{6}$ in the second row and $\mathbf{u}_{6}$ in the last row of Fig.~\ref{fig:fig2}.
Usually, these tiny regions are less likely to be the salient regions we search for.
Besides, these tiny regions often have been captured by other leading eigenvectors as well. Therefore, such eigenvectors usually have low discriminability and may even worsen the final results by overemphasizing tiny regions.Therefore, we evaluate the discriminability of eigenvector $\mathbf{u}_{l}$ by its variance $var(\mathbf{u}_{l})$, and filter out eigenvectors with variance values below a threshold, $v$.

\subsubsection{Normalization}\label{sect:normalization}
After the above local enhancement operations, each original diffusion matrix $\mathbf{A}_i^{-1}$ becomes $\bar{\mathbf{A}}_i^{-1}= \bar{\mathbf{U}}_i \bar{\mathbf{\Lambda}_i}^{-1}\bar{\mathbf{U}}_i^T$ ($1 \leq i \leq M$) in Eq.~\ref{eq:newdiffu}.  Immediately, we may compute $\mathbf{\bar{H}}(i)$ of the $i$-th ($1 \leq i \leq L$) training sample using its enhanced diffusion matrices to replace $\mathbf{H}(i)$ in Eq.~\ref{eq:w} and obtain $\mathbf{w}$. However, this usually is problematic as the saliency vectors computed on different samples and/or by different matrix-seed combinations often exhibit inconsistent ranges of componential values. Therefore, in order to derive an optimal $\mathbf{w}$ of generic applicability, we first normalize the saliency vector of each sample computed by each matrix-seed combination, as explained below.

On each specific image sample, for each matrix-seed combination $\left( \bar{\mathbf{A}_{i}}^{-1}, \mathbf{s}^j \right)$, $1 \leq i \leq M$, $1 \leq j \leq K$, we need to normalize the saliency vector $\bar{\mathbf{y}}^{i,j} = \bar{\mathbf{A}}_{i}^{-1} \mathbf{s}^j = \bar{\mathbf{U}}_i \bar{\mathbf{\Lambda}}_i^{-1} \bar{\mathbf{U}}_i^T \mathbf{s}^j$ to range its componential values to $[0,1]$.  It is commonly known that a vector $\mathbf{x}$ whose componential values extend a range of $[p,q]$ may be normalized by
 \begin{equation}
 \begin{aligned}
    \hat{\mathbf{x}}&=b\mathbf{1}+\frac{\mathbf{x}}{a},\\
    b&=\frac{p}{p-q},\\
    a&=-(p-q).
 \end{aligned}
 \label{eq:ab}
 \end{equation}
Similarly, we normalize $\bar{\mathbf{y}}^{i,j}$ with a componential value range of $[p,q]$ by

\begin{equation}
 \begin{aligned}
    \hat{\mathbf{y}}^{i,j}&=[\mathbf{u}_1, \mathbf{\bar{U}}_i]\left[
	\begin{array}{cc}
	\lambda'{_1}^{-1}&\mathbf{0}^{T}\\
	\mathbf{0}&\frac{1}{\hat{a}}\bar{\mathbf{\Lambda}}_i^{-1}\\
	\end{array}
	\right][\mathbf{u}_1, \bar{\mathbf{U}}_i]^T \mathbf{s}^j\\
	&=\lambda'{_1}^{-1}\mathbf{u}_1\mathbf{u}_1^{T}\mathbf{s}^j + \frac{1}{\hat{a}}\bar{\mathbf{U}}_i\bar{\mathbf{\Lambda}}_i^{-1}\bar{\mathbf{U}}_i^T\mathbf{s}^j\\
	&=\lambda'{_1}^{-1}\mathbf{u}_1\mathbf{u}_1^{T}\mathbf{s}^j + \frac{\bar{\mathbf{y}}^{i,j}}{\hat{a}},
 \end{aligned}
 \label{eq:norm}
 \end{equation}
where $\mathbf{u}_1$ is the constant vector and $\lambda'{_1}$ and $\hat{a}$ are scalars to be determined. By comparing Eq.s~\ref{eq:ab} and~\ref{eq:norm}, we set $\hat{a}=a$ and $\lambda'{_1}^{-1}\mathbf{u}_1\mathbf{u}_1^{T}\mathbf{s}^j=b\mathbf{1}$ for the normalization. Equivalently, we have $\hat{a}=-(p-q)$ and $\lambda'{_1}=\sum_{i=1}^{N}\mathbf{{s}}^j(i)(p-q)/p$.

In essence, the above normalization process refines $\bar{\mathbf{A}}_i^{-1}$ to
\begin{equation}
 \begin{aligned}
    \hat{\mathbf{A}}_{i,j}^{-1}&=\hat{\mathbf{U}}_i\hat{\mathbf{\Lambda}}_i^{-1}\hat{\mathbf{U}}_i^T\\
             &=[\mathbf{u}_1, \bar{\mathbf{U}}_i]\left[
	\begin{array}{cc}
	\lambda'{_1}^{-1}&\mathbf{0}^{T}\\
	\mathbf{0}&\frac{1}{\hat{a}}\bar{\mathbf{\Lambda}}_i^{-1}\\
	\end{array}
	\right][\mathbf{u}_1, \bar{\mathbf{U}}_i]^T
 \end{aligned}
 \label{eq:final}
 \end{equation}
which is used as the final diffusion matrix for $\mathbf{s}^j$ on the specific image sample.

Using the finally refined diffusion matrices, we compute $\hat{\mathbf{H}}(i)$ for the $i$-th ($1 \leq i \leq L$) training sample to replace $\mathbf{H}(i)$ in Eq.~\ref{eq:w} and finally obtain the solution of $\mathbf{w}$.

\subsection{Choice of Seeds} \label{sect:choiceofseeds}

We utilize the foreground and background prior to design two kinds of seed vectors and use them as $\mathbf{s}^j$, $1 \leq j \leq K$ in Eq.s~\ref{eq:superseed} and~\ref{eq:newdiffu}.

Firstly, we assume that nodes closer to the center of image are more salient, and initialize a sequence of Gaussian-filter-like images (with different variances) to compute the first kind of seed vectors, as people usually put salient objects in the central foreground area when taking a photo.

Secondly, we assume that nodes located at the border of image are the least salient, and compute the time that other non-border nodes random walk to them to form the seed vector. Nodes that take more time to reach the border nodes are more salient. Note that the transition matrix of random walk can also be derived from the highly discriminative diffusion matrices, $\bar{\mathbf{A}}_i^{-1}$, $1 \leq i \leq M$, as explained in Appendix~\ref{sect:appendix} with an in-depth analysis of the working mechanism of this proposed seed vector construction method.

The foreground and background prior leads to not only good accuracy of seed value estimation, but also high time-efficiency as it avoids an extra pass of color-based preliminary saliency search.

\subsection{Implementation Details}

When constructing the graph, in order to utilize the cross-node correlation in a broader range, we connect not only nodes that are directly adjacent, but also those that are two hops apart. Furthermore, we connect the nodes at the four borders of an image to each other to make a close-loop graph.

The main training steps of the proposed salient object detection algorithm are summarized in Algorithm~\ref{alg:alg2}. As for testing, given an input image $I$, we conduct the superpixel segmentation and graph construction on it and compute its $\hat{\mathbf{H}}$, following the same initialization and local refinement steps in Alg.~\ref{alg:alg2}, and apply the learned weight $\mathbf{w}$ to $\hat{\mathbf{H}}$ to obtain $\hat{\mathbf{y}}=\hat{\mathbf{H}}\mathbf{w}^T$. Finally, we obtain the saliency map $S$ by assigning the value of $\hat{\mathbf{y}}_{i}$ to the corresponding node $v_i$, $1 \leq i \leq N$.

\begin{algorithm} [!t]
    \caption{Super Diffusion (Training)}
    \begin{algorithmic}[1]
    \REQUIRE \quad\\ (a) A list of training images, $[I_1, \cdots, I_L]$,\\
    (b) A list of scale parameters, $[\sigma_1, \cdots, \sigma_m]$,\\
    (c) A list of feature spaces, $[f_1, \cdots, f_n$],\\
    (d) A list of seed computing methods, $[c_1, \cdots, c_K]$,\\
    \hspace{-6mm}\textbf{Initialization:} \quad\\ Segment each training image into $N$ superpixels, use the superpixels as nodes, connect border nodes to each other and connect nodes that are one or two hops away to construct a graph $G$.\\
    \hspace{-6mm}\textbf{Local refinement:} For each training image in (a),\\
    \STATE Compute $\mathbf{A}_i=\mathbf{D}_i^{-1}(\mathbf{D}_i - 0.99\mathbf{W}_i)$ and its eigenvectors $\mathbf{U}_i$ and eigenvalues $\mathbf{\Lambda}_i$ for each setting $i$ in combination of (b)(c);
    \STATE For each $\mathbf{A}_i$, discard the constant eigenvector, the eigenvectors after the eigengap or with low discriminability to get $\bar{\mathbf{A}}_i^{-1}$, $\bar{\mathbf{U}}_i$, $\bar{\mathbf{\Lambda}}_i$ and $\bar{\mathbf{H}}$ by local refinement operations described in Sec.s~\ref{sect:constant} to~\ref{sect:discriminability};
    \STATE For each $\bar{\mathbf{A}}_i^{-1}$ and each seed computation method, $c_j$, in (d),\\
    \begin{enumerate}[label=(\roman*)]
    \item Compute the seed vector, $\mathbf{s}^j$; \\
    \item Re-add the constant eigenvector with an updated eigenvalue, and scale $\bar{\mathbf{\Lambda}}_i$ to normalize $\bar{\mathbf{y}}^{i,j}$ by Eq.~\ref{eq:norm};\\
    \item Correspondingly, re-synthesize $\bar{\mathbf{A}}_i^{-1}$ to get the final diffusion matrix $\hat{\mathbf{A}}_{i,j}^{-1}$ by Eq.~\ref{eq:final}; \\
    \end{enumerate}

   \STATE Integrate all $\hat{\mathbf{y}}^{i,j}$ to get $\hat{\mathbf{H}}$.\\
    \hspace{-6mm}\textbf{Global optimization:} With $\hat{\mathbf{H}}(i)$, $1 \leq i \leq L$, for all the training images computed, \\
    \STATE Substitute $\hat{\mathbf{H}}(i)$ for $\mathbf{H}(i)$ in Eq.~\ref{eq:w} to compute the optimal weight $\mathbf{w}$.
    \ENSURE Weight $\mathbf{w}$.
    \end{algorithmic}
    \label{alg:alg2}
\end{algorithm}

\subsection{Saliency Features as Diffusion Maps}\label{sect:drfi}
Most diffusion-based salient object detection methods (\eg,~\cite{s11,s12,s18,grab,GP,SP,RW,EQCUTS}) rely on raw color features, \eg, they use the mean color vectors of two linked nodes to
compute the edge weight (\cf Eq.~\ref{eq:resp}) and, correspondingly, the affinity matrix and the diffusion matrix. However, the raw color features may sometimes not be well indicative of the saliency. As such, more saliency features have been devised and used by non-diffusion-based salient object detection methods. In particular, the work~\cite{s16} effectively integrates hundreds of saliency features for the task of salient object detection. This has motivated us to integrate more saliency features seamlessly into our super diffusion framework.

By our interpretation of the diffusion mechanism (\cf Sec.~\ref{sect:diffusionmap}), diffusion maps play a key role in saliency computation and nodes with similar (\resp, dissimilar) diffusion maps tend to be assigned similar (\resp, dissimilar) saliency values. Therefore, good diffusion maps themselves should be discriminative which are similar for nodes of similar factual saliency and dissimilar otherwise. As saliency features discriminate salient from non-salient nodes, we use them to construct discriminative maps at the nodes to imitate the diffusion process. We call them diffusion maps as well for the convenience of description.

We denote the $Z$ saliency features by $\mathbf{g}^1, \mathbf{g}^2, \ldots, \mathbf{g}^Z$ with each $\mathbf{g}^i$, $i \in [1, Z]$, being an $N$-dimensional vector containing the corresponding feature values of the nodes. Then we construct a diffusion map for each node by
 \begin{equation}
 \begin{aligned}
 	\mathbf{\Psi}'_i = [\mathbf{1}(i), \mathbf{g}^1(i), \mathbf{g}^2(i), \cdots, \mathbf{g}^Z(i)].
 \end{aligned}
 \label{eq:ftpsi}
 \end{equation}
Incorporating $\mathbf{\Psi}'_i$ into Eq.~\ref{eq:diffmap}, we update $\hat{\mathbf{\Psi}}_{i}$ to
 \begin{equation}
 \begin{aligned}
 	\hat{\mathbf{\Psi}}_{i} = [\alpha_1\mathbf{\Psi}_{i}^1, \alpha_2\mathbf{\Psi}_{i}^2,\cdots,\alpha_{M}\mathbf{\Psi}_{i}^M, \alpha_{M+1}\mathbf{\Psi}_{i}^{M+1}]
 \end{aligned}
 \label{eq:diffmapft}
 \end{equation}
with $\mathbf{\Psi}_{i}^{M+1} = \mathbf{\Psi}'_i$. Correspondingly, we make $\mathbf{A}_{M+1}=\mathbf{U}_{M+1}\mathbf{\Lambda}_{M+1}\mathbf{U}_{M+1}^{T}$ with $\mathbf{U}_{M+1} = [\mathbf{1}, \mathbf{g}^1, \mathbf{g}^2, \cdots, \mathbf{g}^Z]$ and $\mathbf{\Lambda}_{M+1}=diag\{1, 1, \cdots, 1\}$, and update $\mathbf{A}_I^{-1}$ in Eq.~\ref{eq:lapinte} to
 \begin{equation}
 \begin{aligned}
    \mathbf{A}_I^{-1} &= \mathbf{U}_I \mathbf{\Lambda}_I \mathbf{U}_I^T \\
    &= \mathbf{U}_I\left[
	\begin{array}{ccc}
	\alpha_{1}\mathbf{\Lambda}_1^{-1}&\cdots&\cdots\\
	\vdots&\ddots&\vdots\\
	\cdots&\cdots&\alpha_{M+1}\mathbf{\Lambda}_{M+1}^{-1}\\
	\end{array}
	\right]\mathbf{U}_I^T,
 \end{aligned}
 \label{eq:lapinteft}
 \end{equation}
where $\mathbf{U}_I=\left[\mathbf{U}_1, \cdots, \mathbf{U}_{M+1} \right]$.
Further, we update $\mathbf{H}$ and $\mathbf{w}$ by $\mathbf{H}=[\mathbf{y}^{1,1}, \cdots, \mathbf{y}^{1,K}, \mathbf{y}^{2,1}, \cdots, \mathbf{y}^{2,K}, \cdots, \mathbf{y}^{M+1,1}, \cdots, \mathbf{y}^{M+1,K}]$ and $\mathbf{w}=[w_1, w_2, \cdots, w_{(M+1) \times K}]$ for Eq.s~\ref{eq:newdiffu},~\ref{eq:loss} and~\ref{eq:w}.

Finally, for the training and the testing, the procedures described in the previous sections are still conducted except that the steps in Sec.s~\ref{sect:constant}--\ref{sect:discriminability} are not applied on any $\mathbf{A}_{M+1}$ as it is not a common graph-based diffusion matrix. But still, the normalization step in Sec.~\ref{sect:normalization} is conducted for each matrix-seed combination, $\left( \mathbf{A}_{M+1}, \mathbf{s}^j \right)$, $1 \leq j \leq K$, on each specific image sample.

\section{Experiments and Analysis} \label{sect:experiments}
\subsection{Datasets and Evaluation Methods}

Our experiments are conducted on three datasets: the MSRA10K dataset~\cite{s2,s23} with $10K$ images, the DUT-OMRON dataset~\cite{s11} with $5K$ images and the ECSSD dataset~\cite{s14} with $1K$ images. Each image in these datasets is associated with a human-labeled ground truth.

In order to study the performance of our final super diffusion method, we adopt prevalently used evaluation protocols including precision-recall (PR) curves~\cite{s1}, F-measure score which is a weighted harmonic mean of precision and recall~\cite{s1}, mean overlap rate (MOR) score~\cite{s15} and area under ROC curve (AUC) score~\cite{s18}, as described in Sec.~\ref{sect:salientobject}. Further, to analyze how much the local enhancement operations benefit our method, we propose to measure the quality of a diffusion matrix by visual saliency promotion and constrained optimal seed efficiency (COSE), as detailed in Sec.~\ref{sect:promotion} and Sec.~\ref{sect:OSE}, respectively. Finally, in Sec.~\ref{sect:components}, we give an ablation study of all the global and local enhancement operations, to show the effects of different steps in Alg.~\ref{alg:alg2}.

\begin{figure*}[!t]
\centering
\begin{tabular}
{@{\hspace{-1mm}}c@{\hspace{-1mm}} @{\hspace{-1mm}}c@{\hspace{-1mm}}}
    \subfigure[] {
    \label{fig:boost.A}
    \includegraphics[width=0.24\linewidth]{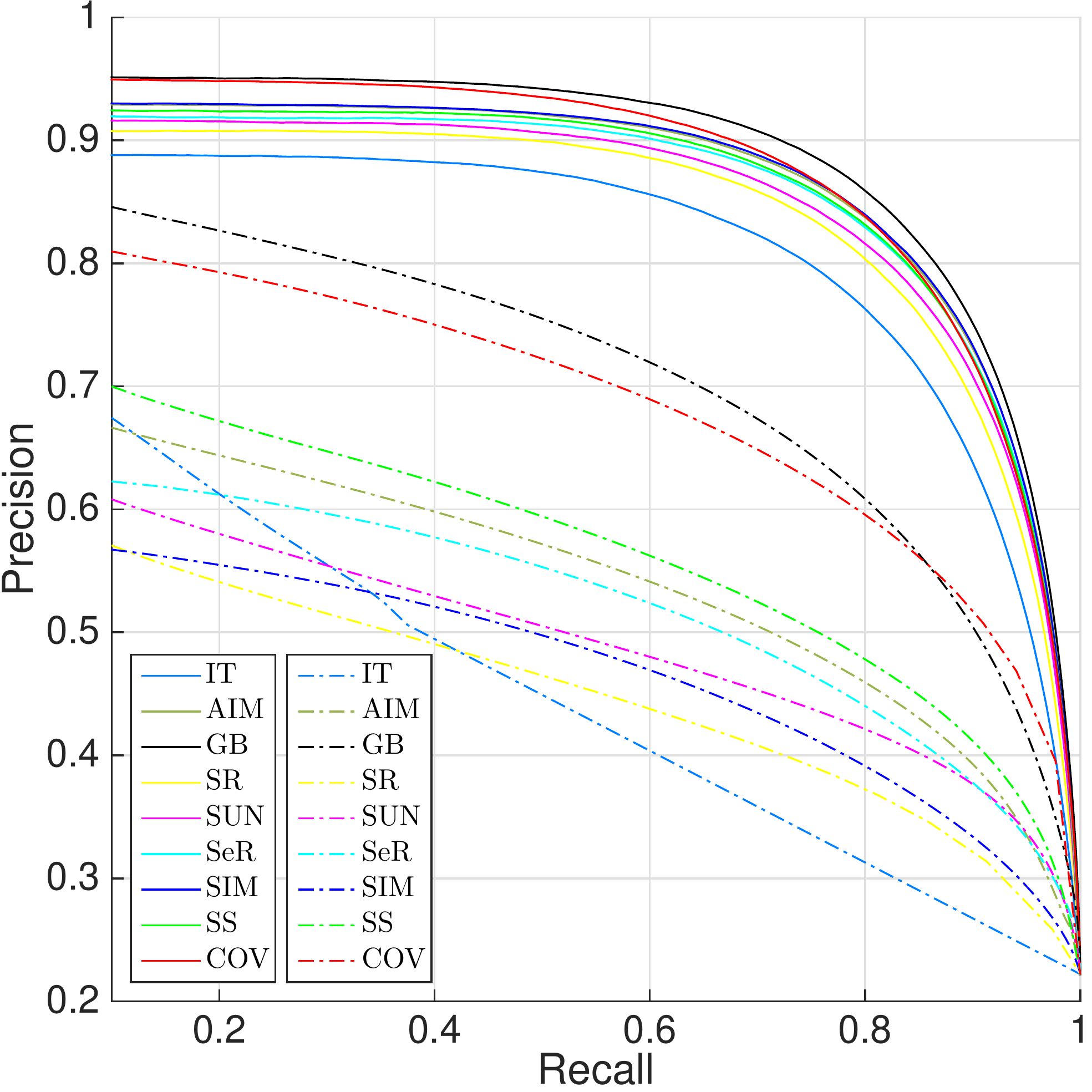}
    }
    \subfigure[] {
    \label{fig:boost.L}
    \includegraphics[width=0.24\linewidth]{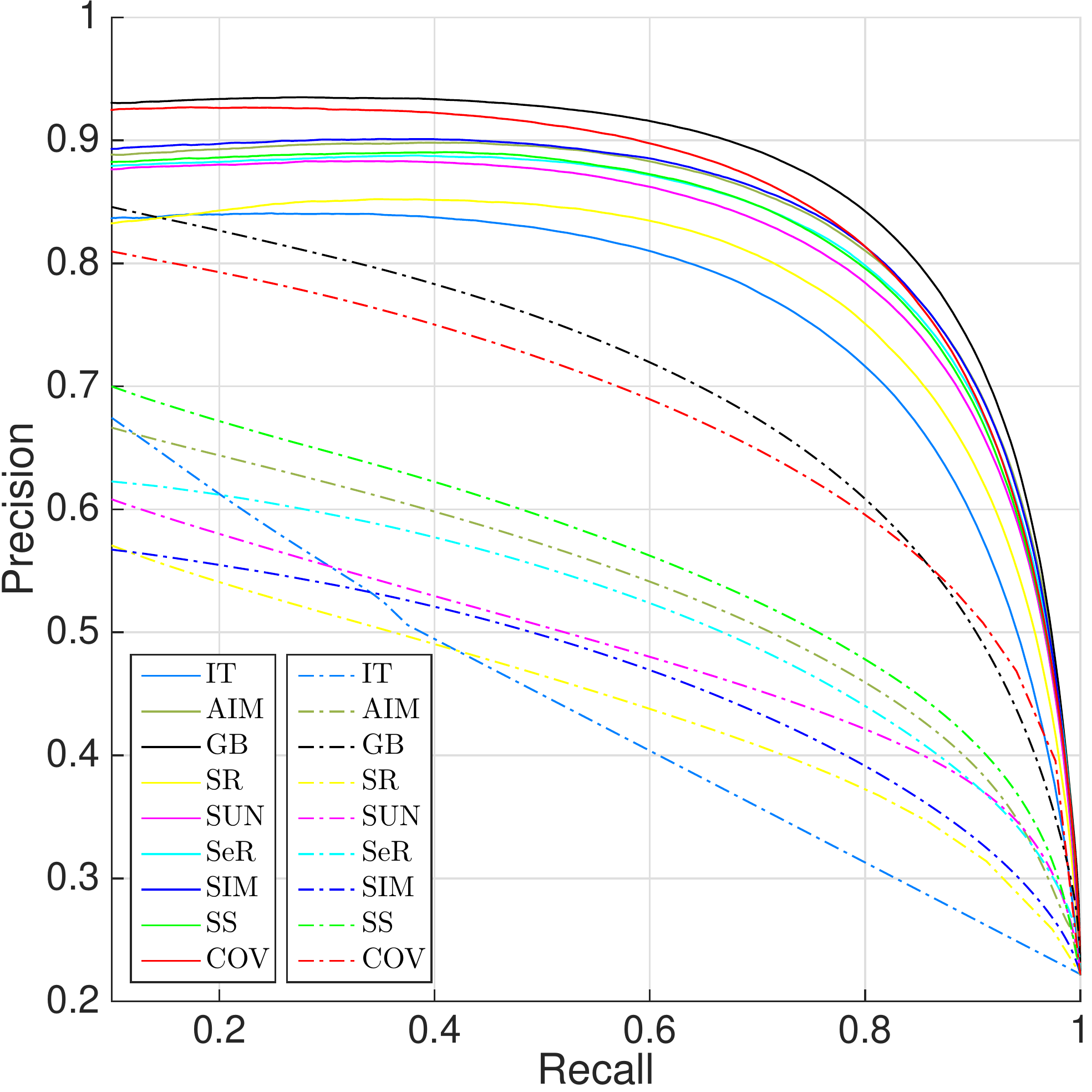}
    }
    \subfigure[] {
    \label{fig:boost.Lrw}
    \includegraphics[width=0.24\linewidth]{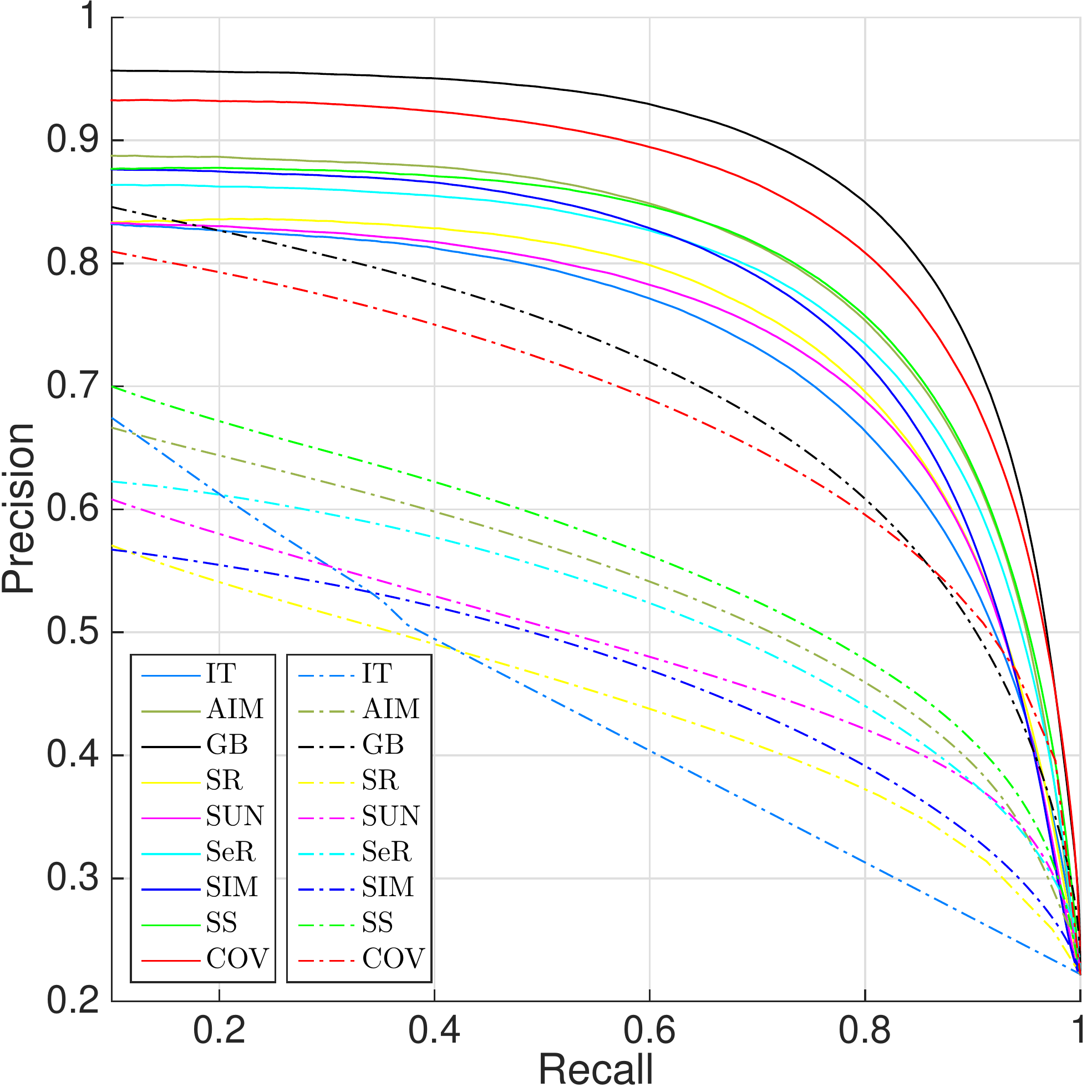}
    }
    \subfigure[] {
    \label{fig:boost.spres}
    \includegraphics[width=0.24\linewidth]{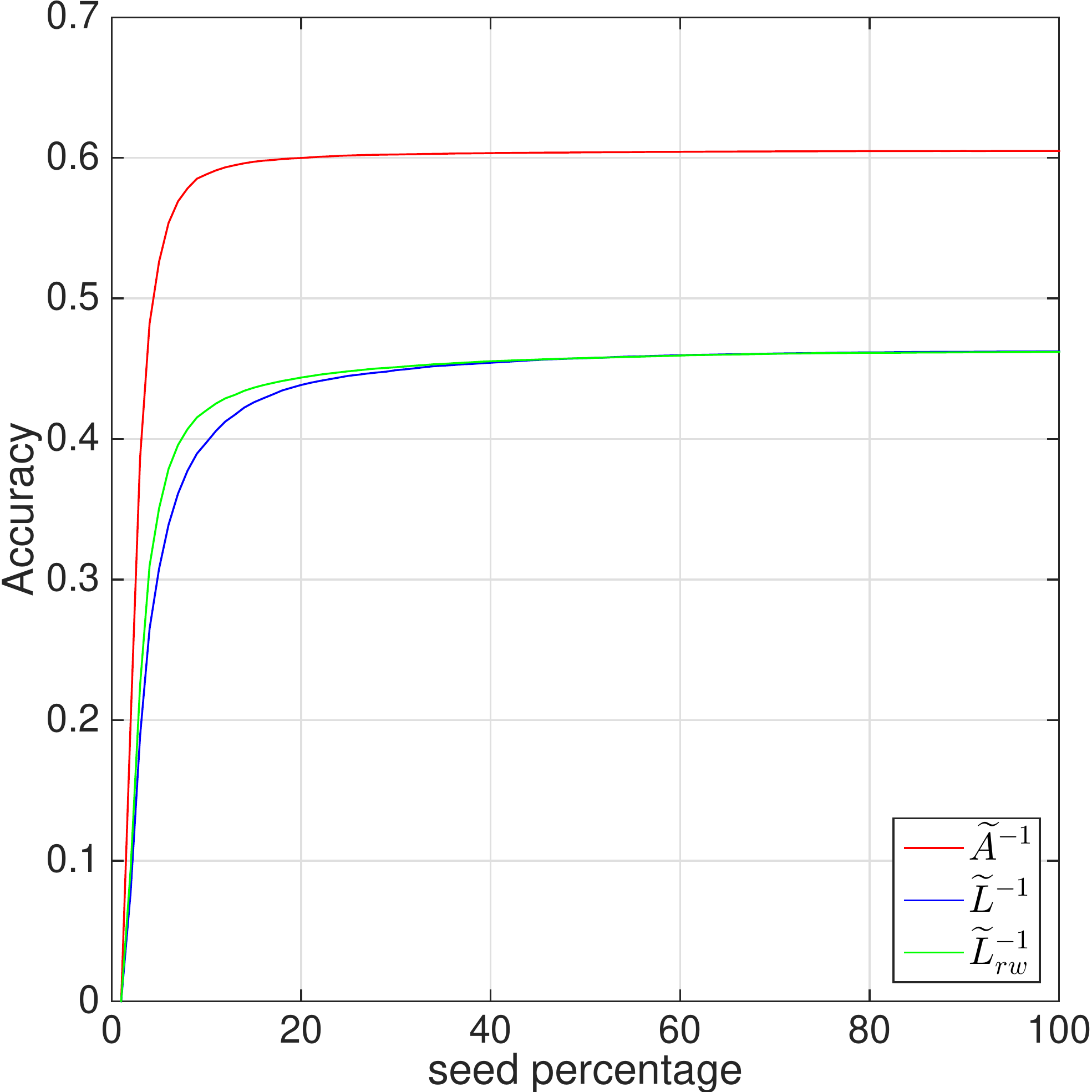}
    }
\end{tabular}
\caption{PR curves of nine visual saliency detection methods before (dash line) and after (solid line) diffusion by (a) $\hat{\mathbf{A}}_{1}^{-1}$, (b) $\tilde{\mathbf{L}}^{-1}$, and (c) $\tilde{\mathbf{L_{rw}}}^{-1}$. The constrained optimal seed efficiency curves for $\hat{\mathbf{A}}_{1}^{-1}$, $\tilde{\mathbf{L}}^{-1}$ and $\tilde{\mathbf{L_{rw}}}^{-1}$ on the MSRA10K dataset are shown in (d).}
\label{fig:boost}
\end{figure*}

\subsection{Experimental Settings}

We choose $11$ different settings for the scale parameter $\sigma$, $\sigma^2 \in{[10, 11,\cdots,20]}$, and 3 different color spaces, $Lab$, $RGB$ and $HSV$, for the feature space, which leads to $11\times3=33$ different diffusion matrices, \ie, $M=33$ for Eq.s~\ref{eq:diffmap},~\ref{eq:lapinte} and~\ref{eq:newdiffu}. We set $v=300$ as the threshold to filter out eigenvectors of low discriminability in Sec.~\ref{sect:discriminability}. For the first kind of seed vectors, we take the Gaussian variance from \{0.5, 1, 2\}. We integrate the saliency features of the work~\cite{s16} into our super diffusion framework (\cf Sec.~\ref{sect:drfi}). For each dataset, we use one half of the images as training samples, and the other half for testing and evaluation.
In Sec.~\ref{sect:promotion} and Sec.~\ref{sect:OSE}, in order to avoid zero eigenvalues, we approximately set $\tilde{\mathbf{L_{rw}}}=\mathbf{D}^{-1}(\mathbf{D}-0.99\mathbf{W})$ and $\tilde{\mathbf{L}}=\mathbf{D}-0.99\mathbf{W}$ when comparing diffusion matrices, as done in the reference~\cite{s11}.
However, our each diffusion matrix $\hat{\mathbf{A}}_{j}$ is directly re-synthesized from $\mathbf{L_{rw}}=\mathbf{D}^{-1}(\mathbf{D}-\mathbf{W})$ by the local refinement.

To comprehensively report the effectiveness of our proposed local refinement operations in Sec.~\ref{sect:measures}, in Sec.~\ref{sect:promotion} and Sec.~\ref{sect:OSE}, we design two experiments to compare the diffusion results with and without the local refinement. In Sec.~\ref{sect:salientobject} and Sec.~\ref{sect:components}, we further demonstrate how much our method gets promoted after global enhancement by integration of diffusions.

\subsection{Promotion of Visual Saliency} \label{sect:promotion}

Visual saliency detection predicts human fixation locations in an image, which are often indicative of salient objects around. Therefore, we use the detected visual saliency as the seed information, and conduct diffusion on it to detect the salient object region in an image. In other words, we promote a visual saliency detection algorithm by diffusion for the task of salient object detection.

In this experiment, we use the results of nine visual saliency detection methods (\ie, IT~\cite{v1}, AIM~\cite{v12}, GB~\cite{v4}, SR~\cite{v13}, SUN~\cite{v14}, SeR~\cite{v15}, SIM~\cite{v16}, SS~\cite{v7} and COV~\cite{v17}) on the MSRA10K dataset as the seed vectors, respectively,
and compare the saliency detection results before and after diffusion. For the diffusion, we test three matrices including $\hat{\mathbf{A}}_{1}^{-1}$, $\tilde{\mathbf{L}}^{-1}$ and $\tilde{\mathbf{L_{rw}}}^{-1}$, which are all computed in $Lab$ feature space with $\sigma^2=10$. It's worth noting that $\hat{\mathbf{A}}_{1}^{-1}$ is only one of our locally refined diffusion matrices (without normalization yet) before the integration.

The PR curves of the nine visual saliency detection methods before and after diffusion by $\hat{\mathbf{A}}_{1}^{-1}$, $\tilde{\mathbf{L}}^{-1}$ and $\tilde{\mathbf{L_{rw}}}^{-1}$ are plotted in Fig.~\ref{fig:boost}(a), (b) and (c), respectively.

Remarkably, as shown in Fig.~\ref{fig:boost}, previous visual saliency detection methods which usually can not highlight the whole salient object all get significantly boosted after diffusion with any of  $\hat{\mathbf{A}}_{1}^{-1}$, $\tilde{\mathbf{L}}^{-1}$ and $\tilde{\mathbf{L_{rw}}}^{-1}$. The promotion is so significant that some promoted methods even outperform some state-of-the-art salient objection detection methods, as observed by comparing Fig.~\ref{fig:boost} and Fig.~\ref{fig:main}. This means that, with a good diffusion matrix, we can fill the performance gap between two branches of saliency detection methods.

Comparing Fig.s~\ref{fig:boost}(a), \ref{fig:boost}(b) and \ref{fig:boost}(c),  we observe that $\hat{\mathbf{A}}_{1}^{-1}$ leads to more significant performance promotion and more consistent promoted performance than $\tilde{\mathbf{L}}^{-1}$ and $\tilde{\mathbf{L_{rw}}}^{-1}$, demonstrating higher effectiveness and robustness of the refined diffusion matrix, $\hat{\mathbf{A}}_{1}^{-1}$, in visual saliency promotion.

 \begin{figure*}
     \hspace{0mm}\includegraphics[width=\linewidth]
     {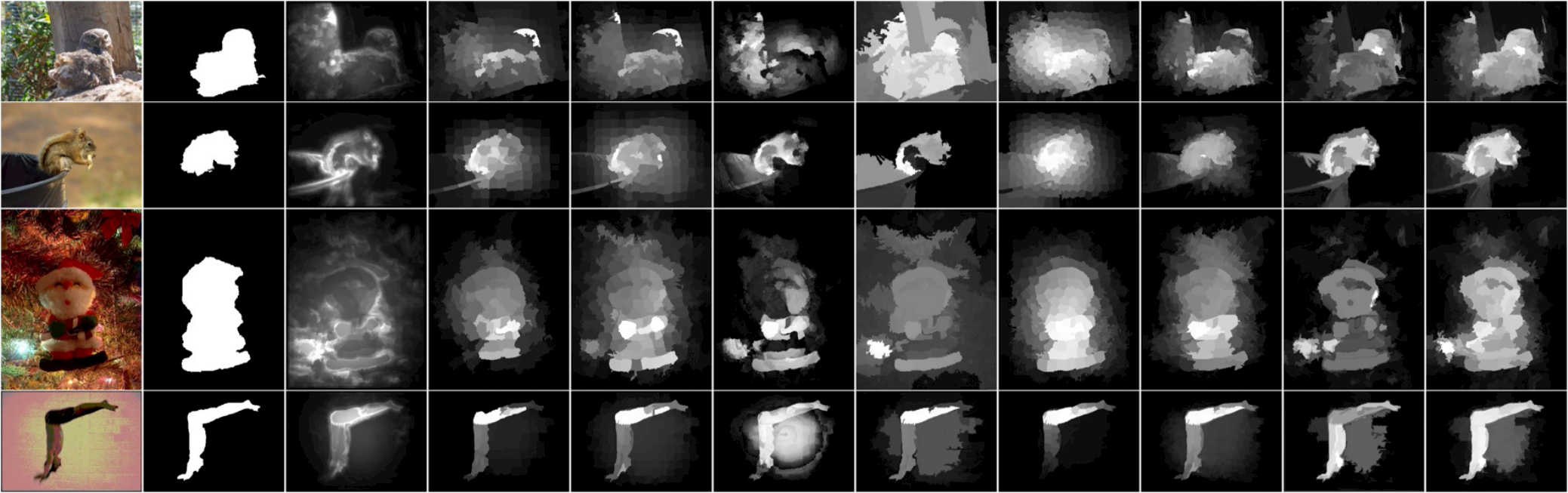}
   \vspace{-2mm}
  {\small\hspace{4mm} SRC  \hspace{10mm} GT \hspace{10mm}  PCA  \hspace{7mm}  GMR  \hspace{8mm}  MC  \hspace{8mm} DSR \hspace{10mm}  HS   \hspace{10mm} GP \hspace{6mm}  Ours(ND) \hspace{4mm} DRFI  \hspace{8mm} Ours}
  \vspace{3mm}
  \caption{Visual comparison of previous approaches to our method and ground truth (GT).}
 \label{fig:final}
 \end{figure*}

\subsection{Constrained Optimal Seed Efficiency} \label{sect:OSE}

We prefer a diffusion matrix to use as little query information or, equally, as few non-zero seed values to derive as close saliency to the ground truth as possible. Correspondingly, for a diffusion matrix, we measure the constrained optimal saliency detection accuracy it may achieve at each non-zero seed value budget, leading to a constrained optimal seed efficiency curve, as detailed below.

Given the ground truth $\mathbf{GT}$ and the diffusion matrix $\mathbf{A}^{-1}$, we hope to find the optimal seed vector, $\mathbf{s}$, that minimizes the residual, $\mathbf{res}$, computed by
 \begin{equation}
 \begin{aligned}
	\mathbf{res}=\mathbf{GT}- \mathbf{A}^{-1}\mathbf{s}.
 \end{aligned}
 \label{eq:eq10}
 \end{equation}
Aiming to reduce the number of non-zero values in $\mathbf{s}$, we turn the residual minimization to a sparse recovery problem, to solve which we adapt the algorithm of orthogonal matching pursuit (OMP)~\cite{a8}, as described in Alg.~\ref{alg:alg3}.

As shown in Alg.~\ref{alg:alg3}, we adapt the residual computation to 	 $\tilde{\mathbf{res}}=\mathbf{GT}-bin(\mathbf{A}^{-1}\mathbf{s})$ in Step $4$, where $bin$ is the binarization operation since $\mathbf{GT}$ is binary; we multiply a factor $\mathbf{GT}(j)$ in Step $1$ to ensure that the non-zero seed values are selected from only the salient region; we solve the nonnegative least-squares problem in Step $3$ to ensure nonnegative elements of $\mathbf{s}$. The adapted OMP will stop when $\|\tilde{\mathbf{res}}\|_2$ is below a threshold, $c$, or the nonnegative seed values at the salient region are all selected, as shown in Step $5$.  We see that the optimization process in Alg.~\ref{alg:alg3} is constrained, \eg, the seeds are selected from only the salient region, the optimization is conducted in a greedy fashion and so forth. Although the saliency detection performance of these resultant seed vectors provides a good reference for our diffusion matrix evaluation, it should be noted that their optimal performance is constrained but not absolute.

In order to obtain the constrained optimal seed efficiency curve over the full range of nonnegative seed value budget, we set $c=0$ in Alg.~\ref{alg:alg3} and, at the $i$-th ($0 \leq i \leq 100$) iteration, we compute and record the pair of nonnegative seed percentage, $r_i$, and saliency detection accuracy, $a_i$, according to the following formulae:
 \begin{equation}
 \begin{aligned}
	r_i&=\frac{100\times\|\mathbf{s}\|_0}{\|\mathbf{GT}\|_0}\%,\\
	a_i&=\frac{\|\mathbf{GT}\|_2-\|\tilde{\mathbf{res}}\|_2}{\|\mathbf{GT}\|_2}.
 \end{aligned}
 \label{eq:accuracy}
 \end{equation}
Based on these ($r_i$, $a_i$) pairs, we can plot the OSE curve of $\mathbf{A}^{-1}$ on an image.

We substitute $\hat{\mathbf{A}}_{1}^{-1}$, $\tilde{\mathbf{L}}^{-1}$ and $\tilde{\mathbf{L}_{rw}}^{-1}$ in last section into Eq.~\ref{eq:eq10} for $\mathbf{A}^{-1}$, respectively. For each diffusion matrix, we plot the average OSE curve over all the images in the MSRA10K dataset, as shown in Fig.~\ref{fig:boost.spres}.  From Fig.~\ref{fig:boost.spres}, we observe that the constrained optimal seed efficiency rises sharply at the beginning and levels off at around the nonnegative seed percentage of $30\%$, that $\hat{\mathbf{A}}_{1}^{-1}$ exhibits significantly higher average constrained optimal seed efficiency than $\tilde{\mathbf{L}}^{-1}$ and $\tilde{\mathbf{L_{rw}}}^{-1}$, and that there is an inherent performance ceiling for each diffusion matrix while $\mathbf{\hat{A}_{1}}^{-1}$ has the highest one. According to the last observation, it appears that the performance of diffusion-based saliency detection is fundamentally determined by the diffusion matrix, again emphasizing the importance in constructing a good diffusion matrix.

 \begin{algorithm} [!t]
    \caption{Adapted Orthogonal Matching Pursuit}
    \begin{algorithmic}[1]
    	\REQUIRE Dictionary$(\mathbf{A}^{-1}_{N{\times}N})$, Signal$(\mathbf{GT}_{N{\times}1})$ and Stop criterion$(c)$\\
	\ENSURE Coefficient vector$(\mathbf{s}_{N{\times}1})$ and Residual$(\mathbf{res})$\\
	\hspace{-6mm}\textbf{Initialization:} $\mathbf{res}=\mathbf{GT}$, $Inds=\emptyset$,\\ $FgInds=\underset{i}{\operatorname{arg}}{\{\mathbf{GT}(i)=1\}}$\\
	\hspace{-6mm}\textbf{Iteration:}
	\STATE $ind=\underset{j}{\operatorname{argmax}}\{|\left<\mathbf{res},\mathbf{A}^{-1}(:,j)\right>|\cdot \mathbf{GT}(j)\}$, $j \in FgInds$;\\
	\STATE $Inds=Inds \cup ind$, $FgInds=FgInds \setminus ind$;
	\STATE $\mathbf{s}(Inds)=\underset{\mathbf{\tilde{s}}\geq0}{\operatorname{argmin}}\|\mathbf{GT}-\mathbf{A}^{-1}(:,Inds)\mathbf{\tilde{s}}\|_2$;
	\STATE $\tilde{\mathbf{res}}=\mathbf{GT}- bin(\mathbf{A}^{-1}\mathbf{s})$,\\
	\IF {$\|\tilde{\mathbf{res}}\|_2\geq c  \wedge FgInds\neq\emptyset $}
		\STATE \textbf{Go to} 1;
	\ENDIF
    \end{algorithmic}
    \label{alg:alg3}
 \end{algorithm}

 \begin{table*}
\centering
\small
\begin{tabular}{|c|c|ccccccccc|}
\hline
 Dataset & Protocol & \multicolumn{1}{c|}{PCA} & \multicolumn{1}{c|}{GMR} & \multicolumn{1}{c|}{MC} & \multicolumn{1}{c|}{DSR} & \multicolumn{1}{c|}{HS} & \multicolumn{1}{c|}{GP} & \multicolumn{1}{c|}{Our(ND)} & \multicolumn{1}{c|}{DRFI} & \multicolumn{1}{c|}{Ours}  \\

 \hline
 \hline

  &Precision & 0.80289                  & \JP{0.89021}                 & \HL{0.89063}                 & 0.8532                   & 0.88492                  & 0.87807                 & 0.8586                    & 0.86753                    & \JLP{0.88712}                                          \\

  \cline{2-2}
 & Recall    & 0.67817                  & 0.752                   & 0.75455                 & 0.73813                  & 0.71551                  & 0.78882                 & \JLP{0.80801}                    & \JP{0.84147}                   & \HL{0.85289}                                           \\
 \cline{2-2}
MSRA10K & F-measure & 0.7702                   & 0.85399                 & 0.85504                 & 0.82357                  & 0.83908                  & \JLP{0.85573}                 & 0.84637                    & \JP{0.86138}                    & \HL{0.87898}                                          \\
\cline{2-2}
&AUC       & 0.94111                  & 0.94379                 & 0.95074                 & 0.95888                  & 0.93264                  & 0.96358                 & \JLP{0.9651}                    & \JP{0.97758}                   & \HL{0.98252}                                          \\ \cline{2-2}
&Overlap   & 0.57652                  & 0.69254                 & 0.69386                 & 0.65398                  & 0.65576                  & 0.71011                 & \JLP{0.71053}                     & \JP{0.73879}                   & \HL{0.76318}                                         \\

\hline
\hline

& Precision & 0.66047                   & 0.76865                 & \JLP{0.77004}                 & 0.74891                  & 0.76924                  & 0.7376                 & 0.74345                    & \JP{0.77903}                    & \HL{0.7983}                                            \\ \cline{2-2}
&Recall    & 0.52427                  & 0.64498                 & 0.65227                 & 0.64544                  & 0.53912                  & 0.68775                 & \JLP{0.69936}                    & \JP{0.7259}                   & \HL{0.73749}                                           \\ \cline{2-2}
ECSSD& F-measure & 0.62311                  & 0.73608                 & \JLP{0.73924}                 & 0.72219                   & 0.70027                  & 0.72547                 & 0.73279                    & \JP{0.76609}                    & \HL{0.78339}                                           \\ \cline{2-2}
&AUC       & 0.87643                  & 0.89127                 & 0.91113                 & 0.9154                  & 0.88534                  & 0.91663                 & \JLP{0.92349}                   & \JP{0.94521}                   & \HL{0.95339}                                           \\ \cline{2-2}
&Overlap   & 0.39517                  & 0.52335                 & 0.53065                 & 0.51352                  & 0.45799                  & 0.53146                 & \JLP{0.5423}                   & \JP{0.58415}                   & \HL{0.60073}                                            \\

\hline
\hline

& Precision & 0.4784                    & \JLP{0.55641}                 & 0.54751                 & 0.53419                  & \HL{0.57452}                  & 0.52623                 & 0.51532                    & 0.54588                   & \JP{0.56297}                                            \\ \cline{2-2}
&Recall    & 0.63225                  & 0.62502                 & 0.67161                 & 0.68049                  & 0.57653                  & 0.65185                 & \JLP{0.68299}                    & \JP{0.76822}                   & \HL{0.77057}                                           \\ \cline{2-2}
DUT-OMRON& F-measure & 0.50686                  & 0.57087                 & 0.57189                 & 0.56208                   & \JLP{0.57498}                  & 0.55072                 & 0.54627                    & \JP{0.58495}                   & \HL{0.60029}                                           \\ \cline{2-2}
&AUC       & 0.88716                  & 0.85276                 & 0.88691                 & \JLP{0.89901}                  & 0.8607                  & 0.86891                 & 0.88407                   & \JP{0.93353}                   & \HL{0.9355}                                           \\ \cline{2-2}
&Overlap   & 0.34133                  & 0.42156                 & \JLP{0.42529}                 & 0.40828                  & 0.39755                  & 0.41029                 & 0.40978                   & \JP{0.45105}                   & \HL{0.46575}                                            \\

\hline

\end{tabular}

\vspace{3mm}
\caption{
Performance statistics of different algorithms on the five protocols and the two datasets. For each dataset and protocol, the top three results are highlighted in red, blue and green, respectively.}
\label{tab:table1}
\end{table*}

\begin{figure*}
\centering
\begin{tabular}
{@{\hspace{-1.2mm}}c@{\hspace{-1.2mm}} @{\hspace{-1.2mm}}c@{\hspace{-1.2mm}}}
    \subfigure[]{
    \label{fig:main.msra}
    \includegraphics[width=0.24\linewidth]{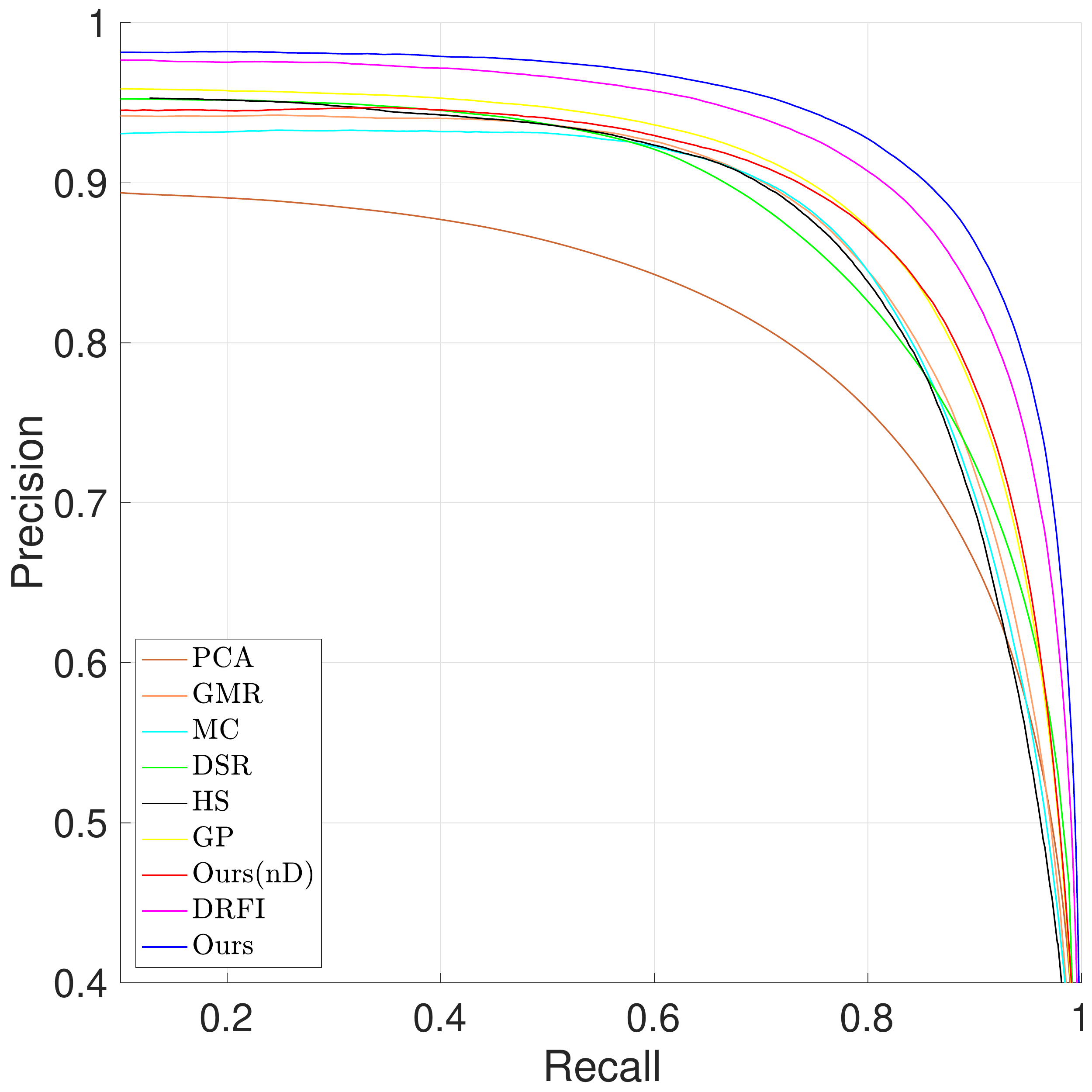}
    }
    \subfigure[]{
    \label{fig:main.ecssd}
    \includegraphics[width=0.24\linewidth]{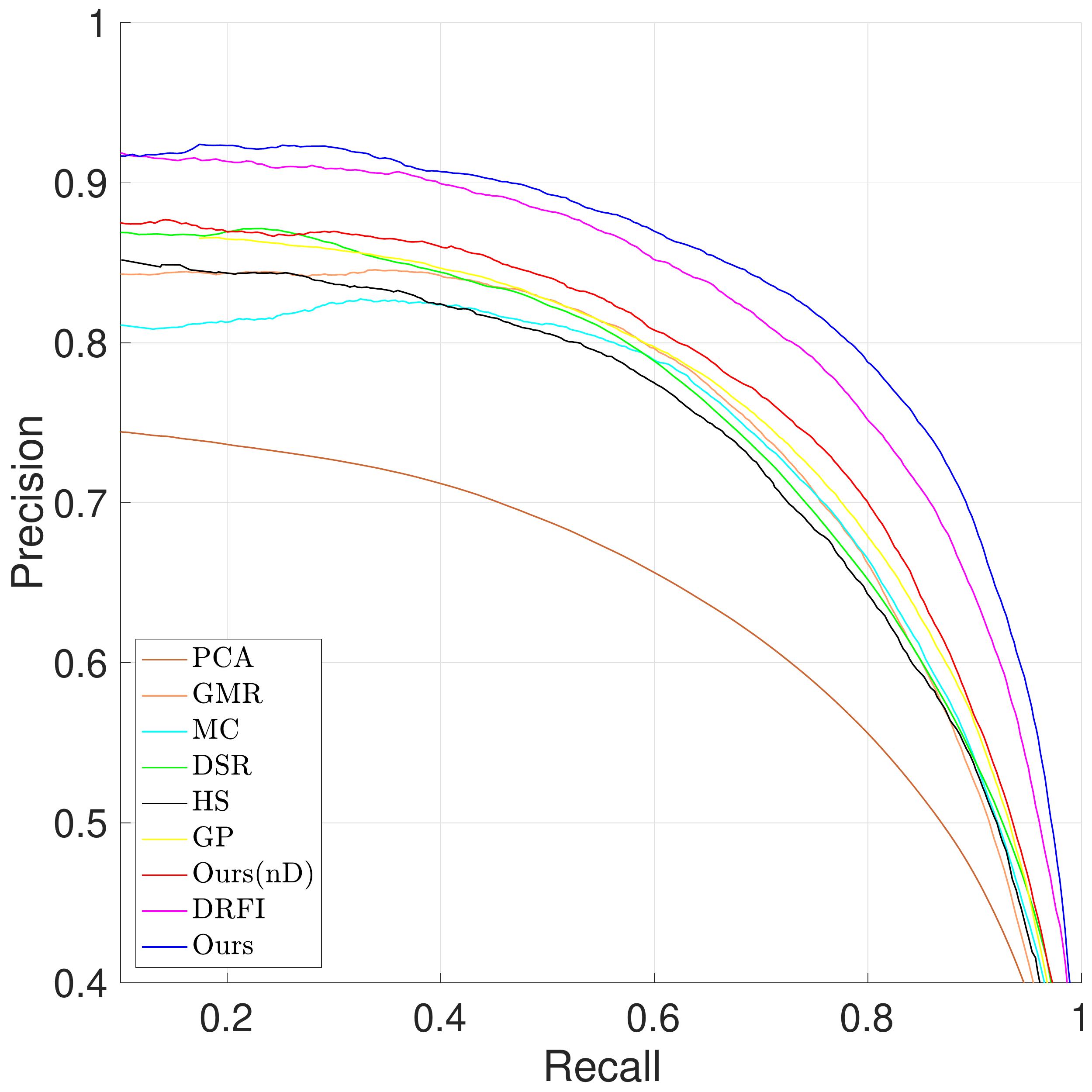}
    }
    \subfigure[]{
    \label{fig:main.dut}
    \includegraphics[width=0.24\linewidth]{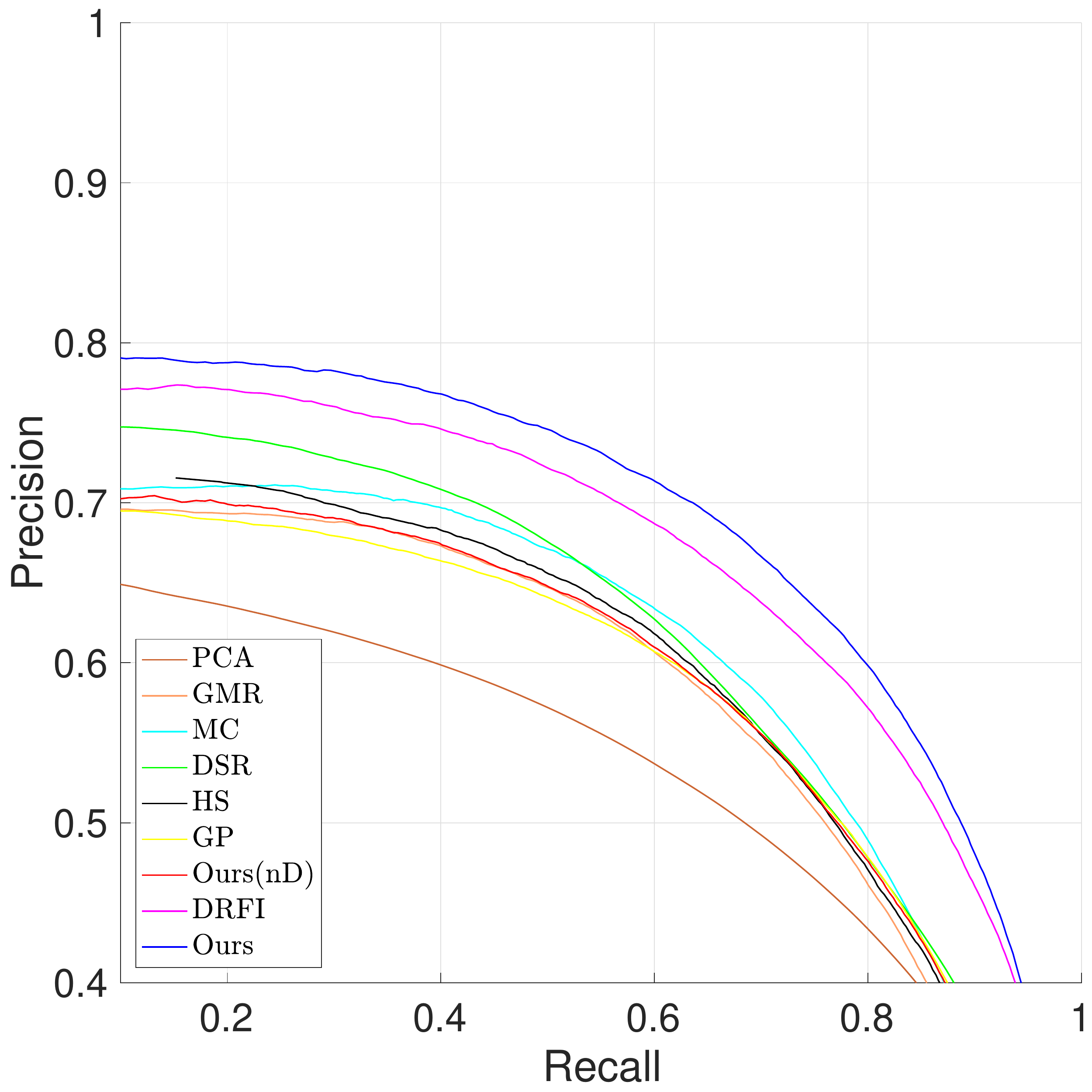}
    }
    \subfigure[]{
    \label{fig:main.steps}
    \includegraphics[width=0.24\linewidth]{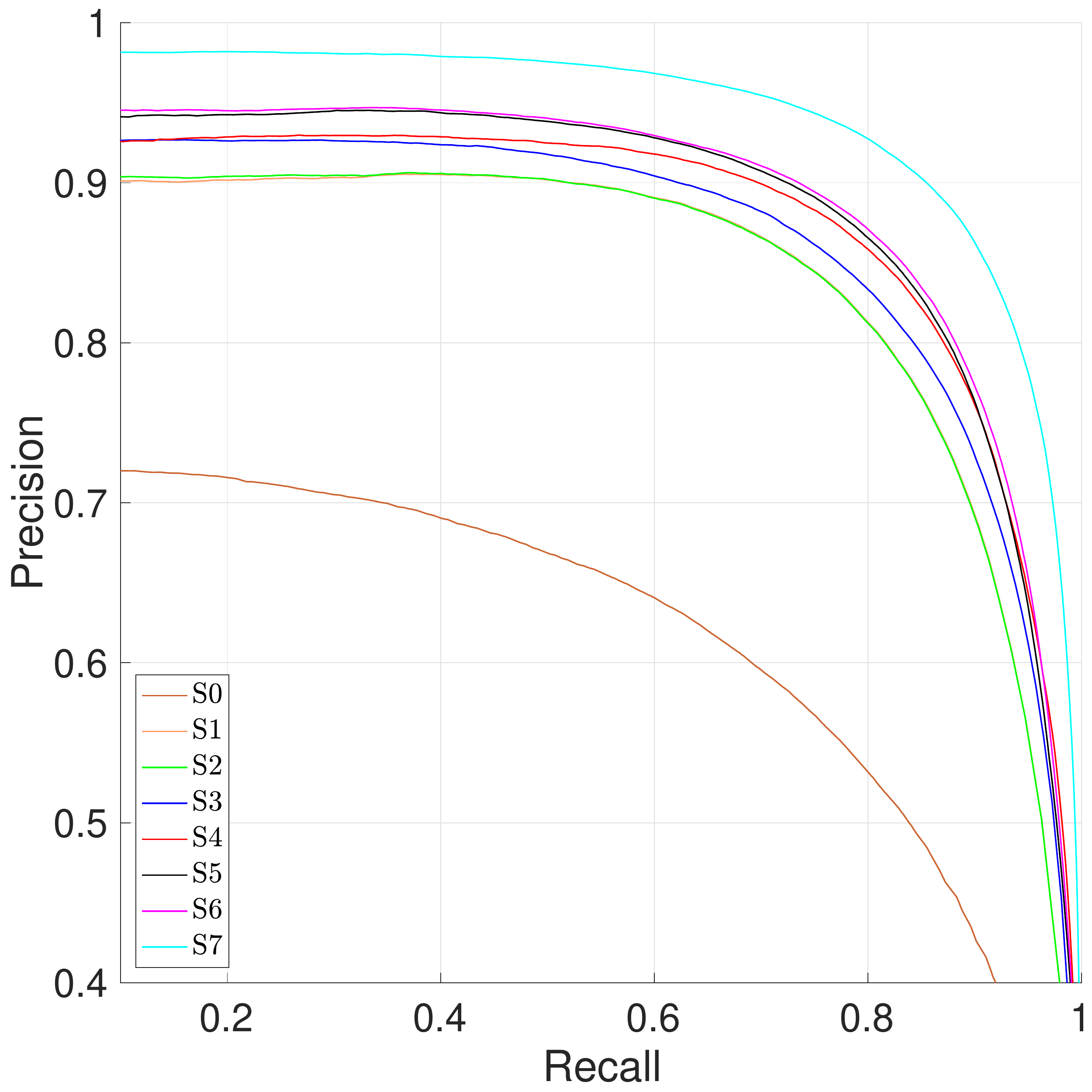}
    }
\end{tabular}
\caption{PR curves for all the algorithms on (a) the MSRA10K dataset~\cite{s2,s23}, (b) the ECSSD dataset~\cite{s14}, (c) the DUT-OMRON dataset~\cite{s11}, and
(d) PR curves for building steps of our scheme on the MSRA10K dataset~\cite{s2,s23}.
}
\vspace{-2mm}
\label{fig:main}
\end{figure*}

\subsection{Salient Object Detection}\label{sect:salientobject}

We experimentally compare our methods (Ours and Ours(ND)) with eight other recently proposed ones including PCA~\cite{s31}, GMR~\cite{s11}, MC~\cite{s12}, DSR~\cite{s28}, HS~\cite{s14}, GP~\cite{GP} and DRFI~\cite{s16} on salient object detection. When evaluating these methods, we either use the results from the original authors (when available) or run our own implementations. Among these methods, GMR, MC, and GP are the diffusion-based methods which lead to outstanding performance, and DRFI is the approach that integrates hundreds of saliency features and yields top performance on the saliency benchmark study~\cite{a10}\footnote{We have noted that recently proposed CNN based methods such as~\cite{deep1,deep2} achieved the best performance on their own report. However, these methods are hard to re-implement and saliency benchmark study~\cite{a10} does not list these works either.}. Ours(ND) (\resp, Ours) is our super diffusion method without (\resp, with) the saliency features of DRFI~\cite{s16} integrated. 
 
We plot the PR curves of all the nine methods on MSRA10K dataset, ECSSD dataset and DUT-OMRON dataset in Fig.s~\ref{fig:main.msra}, ~\ref{fig:main.ecssd} and ~\ref{fig:main.dut}, respectively. Further, we provide the performance statistics on the five prevalent protocols for all the methods on the three datasets in Tab.~\ref{tab:table1}. From both Fig.~\ref{fig:main} and Tab.~\ref{tab:table1}, we clearly observe that Ours(ND) outperforms the other diffusion-based methods and, after integrating the saliency feature of DRFI, Ours yields the top performance.

For visual comparison, we show in Fig.~\ref{fig:final} the saliency object detection results by the benchmark methods and our methods on several images in MSRA10K. From Fig.~\ref{fig:final}, we observe clearly that Ours produces much closer results to the ground truth than the others. It is worth noting that, although Ours(ND) and DRFI both miss some salient regions, after integrating the saliency features of DRFI into the super diffusion framework, Ours successfully highlights most of the salient regions.

\subsection{Effects of Building Steps}\label{sect:components}

In this section, we demonstrate the incremental effects of building steps in the proposed global and local enhancement operations (\cf Sec.~\ref{sect:integration}, Sec.~\ref{sect:measures} and Sec.~\ref{sect:drfi}), as detailed below.

For each test image, we may obtain eight PR curves, $S0$ to $S7$. We start from its $S0$ and progressively obtain $S1$ to $S7$ when the constant eigenvector is discarded, the eigenvectors after the eigengap are filtered out, the discriminability weighting is conducted, diffusion maps derived from multiple color spaces are integrated, diffusion maps derived from multiple scales are integrated, multiple diffusion seeds are integrated and saliency features are imported as diffusion maps, respectively. Experimenting on the whole MSRA10K dataset~\cite{s2,s23}, we obtain the average PR curves for $S0$ to $S7$, as plotted in Fig.~\ref{fig:main.steps}.

From Fig.~\ref{fig:main.steps}, we observe that all the local and global enhancement operations consistently improve the performance, and the introduction of saliency features as diffusion maps leads to the top performance.

\section{Conclusions}

In this work, we have proposed a super diffusion framework that systematically integrates various diffusion matrices, saliency features and seed vectors into a generalized diffusion system for salient object detection. To the best of our knowledge, this is the first framework of this kind ever published.  The whole framework is theoretically based on our novel re-interpretation of the working mechanism of diffusion-based salient object detection, \ie, diffusion maps are core functional elements and the diffusion process is closely related to spectral clustering in general. It takes a learning-based approach and provides a closed-form best solution to the global weighting for the integration. At the local level, it refines each diffusion matrix by getting rid of less discriminating eigenvectors, normalizes each specific saliency vector, and even incorporates discriminative saliency features as diffusion maps.   As a result, the proposed framework produces a highly robust salient object detection scheme, yielding the state-of-the-art performance.

It is worthwhile to emphasize that the proposed super diffusion framework is open and extensible.  Besides those employed in this work, it may integrate any other diffusion matrices, saliency features and/or seed vectors as well into the system specifically trained for any application with specific criterion in saliency object detection. In particular, it would be interesting to integrate various CNN-learned saliency features (\eg,~\cite{deep1,deep2}.) into the proposed framework and examine the performance promotion for specific applications. This is being planned for our future work.

\begin{appendices}
\section{}\label{sect:appendix}

In this appendix, we give the proof of Eq.~\ref{eq:eq4} and clarify the working mechanism of the second kind of seed vectors proposed in Sec.~\ref{sect:choiceofseeds}.

\subsection{Proof of Eq.~\ref{eq:eq4}}

The work~\cite{s12} duplicates the superpixels around the image borders as virtual background absorbing nodes, and sets the inner nodes as transient nodes, thus constructing an Absorbing Markov Chain.  It computes the absorbed time for each node as its saliency value. In Eq.s 1 and 8 in the work~\cite{s12}, it formulates the transition matrix as
 \begin{equation}
 \begin{aligned}
	 &\mathbf{P} = \mathbf{D}^{-1}\mathbf{W}=
	 \begin{pmatrix}
 	\mathbf{Q} ~~~ \mathbf{R}\\
	\mathbf{0} ~~~ \mathbf{I}
	 \end{pmatrix},~ \\
 \end{aligned}
 \label{eq:first}
 \end{equation}
where the first $m$ nodes are transient nodes and the last $N-m$ nodes are absorbing nodes, $\mathbf{Q}\in[0, 1]^{m \times m}$ contains the transition probabilities between any pair of transient nodes, while $\mathbf{R}\in[0, 1]^{m \times (N-m)}$ contains the probabilities of moving from any transient node to any absorbing node. $0$ is the $(N-m) \times m$ zero matrix and $\mathbf{I}$ is the $(N-m) \times (N-m)$ identity matrix. According to Eq. 2 in the work~\cite{s12}, the absorbed time for $m$ transient nodes is
 \begin{equation}
 \begin{aligned}
	&\mathbf{y^*} = (\mathbf{I}-\mathbf{Q})^{-1}\mathbf{c},\\
 \end{aligned}
 \label{eq:N}
 \end{equation}	
where $\mathbf{c}$ is a $m$ dimensional column vector all of whose elements are $1$.

In our derivation, we extend Eq.~\ref{eq:N} to
 \begin{equation}
 \begin{aligned}
	&\mathbf{y^*} = (\mathbf{I}-\mathbf{Q})^{-1}\mathbf{c} = (\mathbf{Q}^0+\mathbf{Q}^1+\mathbf{Q}^2+\ldots)\mathbf{c}.\\
 \end{aligned}
 \label{eq:N1}
 \end{equation}	
and compute the $n$-th power of $\mathbf{P}$ as
 \begin{equation}
 \begin{aligned}
	&\mathbf{P}^n=
	 \begin{pmatrix}
 	 &\mathbf{Q}^n~~~&(\mathbf{Q}^0+\mathbf{Q}^1+\ldots+\mathbf{Q}^{n-1})\mathbf{R}\\
	 &\mathbf{0}~~~&\mathbf{I}\\
	 \end{pmatrix}.
 \end{aligned}
 \label{eq:P}
 \end{equation}
As the absorbed time for absorbing nodes is $0$, we define the absorbed time for all the nodes as $\mathbf{y} =
	\begin{pmatrix}
 	 \mathbf{y^*}\\
	 \mathbf{0}
	 \end{pmatrix}
$.  From Eq.s~\ref{eq:N1},~\ref{eq:P} and~\ref{eq:first}, we have
 \begin{equation}
 \begin{aligned}
	 \mathbf{y} &= (\mathbf{P}^0+\mathbf{P}^1+\mathbf{P}^2+\ldots)\mathbf{x} = (\mathbf{1}-\mathbf{P})^{-1}\mathbf{x}\\&=(\mathbf{D}^{-1}(\mathbf{D}-\mathbf{W}))^{-1}\mathbf{x}=\mathbf{L_{rw}}^{-1}\mathbf{x},\\
 \end{aligned}
 \label{eq:aeq5}
 \end{equation}
where $\mathbf{x}=	\begin{pmatrix}
 	 \mathbf{c}\\
	 \mathbf{0}
	 \end{pmatrix}$.  This completes the proof of Eq.~\ref{eq:eq4}.

Further, based on our re-interpolation of the diffusion (\emph{ref.} Sec.~\ref{sect:diffusionmap}),
 \begin{equation}
 \begin{aligned}
	 \mathbf{y}_i & = \sum_{j=1}^N \mathbf{x}_j\left<\mathbf{\Psi_{{L_{rw}}_i}}, \mathbf{\Psi_{{L_{rw}}_j}}\right>\\
	 & = \sum_{j=1}^m\left<\mathbf{\Psi_{{L_{rw}}_i}}, \mathbf{\Psi_{{L_{rw}}_j}}\right>,
 \end{aligned}
 \label{eq:abs_re}
 \end{equation}
meaning that the absorbed time of each node is equal to the sum of the inner products of its diffusion map with those of all the $m$ non-border nodes on the Absorbing Markov Chain.

\subsection{The Second Kind of Seed Vectors}

In effect, after connecting all the nodes at the four borders of the image,
we have constructed a graph similar to the Absorbing Markov Chain.
For every node at the border, it connects with all $bn$ border nodes (including itself) and only $bm$ non-border nodes ($bm \ll bn$), meaning that once a random walk reaches a border node, it will less likely escape from the border node set. Therefore, we may assume that all the non-border nodes are transient nodes and all the border nodes are background absorbing nodes.

Accordingly, we compute the absorbed time of all nodes in an image to form a seed vector of the second kind,
That is, following Eq.~\ref{eq:abs_re}, we have
\begin{equation}
  \mathbf{e}^i(j) = \sum_{k=1}^d \left<\bar{\mathbf{\Psi}}_{j}^i, \bar{\mathbf{\Psi}}_{k}^i\right>
\label{eq:eq11-0}
\end{equation}
or, equivalently,	
\begin{equation}
\mathbf{e}^i = \bar{\mathbf{A}}_i^{-1}\mathbf{z}
\label{eq:eq11}
\end{equation}
where $\mathbf{z}(k) = 1$ if $v_k$ is a non-border node and $\mathbf{z}(k) = 0$ otherwise.

\end{appendices}

\ifCLASSOPTIONcaptionsoff
  \newpage
\fi



%


\bibliographystyle{IEEEtran}
\bibliography{tip.bib}

\begin{thebibliography}{10}
\providecommand{\url}[1]{#1}
\csname url@samestyle\endcsname
\providecommand{\newblock}{\relax}
\providecommand{\bibinfo}[2]{#2}
\providecommand{\BIBentrySTDinterwordspacing}{\spaceskip=0pt\relax}
\providecommand{\BIBentryALTinterwordstretchfactor}{4}
\providecommand{\BIBentryALTinterwordspacing}{\spaceskip=\fontdimen2\font plus
\BIBentryALTinterwordstretchfactor\fontdimen3\font minus
  \fontdimen4\font\relax}
\providecommand{\BIBforeignlanguage}[2]{{%
\expandafter\ifx\csname l@#1\endcsname\relax
\typeout{** WARNING: IEEEtran.bst: No hyphenation pattern has been}%
\typeout{** loaded for the language `#1'. Using the pattern for}%
\typeout{** the default language instead.}%
\else
\language=\csname l@#1\endcsname
\fi
#2}}
\providecommand{\BIBdecl}{\relax}
\BIBdecl

\bibitem{v1}
L.~Itti, C.~Koch, and E.~Niebur, ``A model of saliency-based visual attention
  for rapid scene analysis,'' \emph{IEEE PAMI}, 1998.

\bibitem{v4}
J.~Harel, C.~Koch, and P.~Perona., ``Graph-based visual saliency,''
  \emph{NIPS}, 2006.

\bibitem{v5}
T.~Judd, K.~Ehinger, F.~Durand, and A.~Torralba, ``Learning to predict where
  humans look,'' \emph{ICCV}, 2009.

\bibitem{v7}
X.~Hou, J.~Harel, and C.~Koch, ``Image signature: Highlighting sparse salient
  regions,'' \emph{IEEE PAMI}, 2012.

\bibitem{v8}
D.~Gao and N.~Vasconcelos, ``Decision-theoretic saliency: Computational
  principles, biological plausibility, and implications for neurophysiology and
  psychophysics,'' \emph{Neural Computation}, 2009.

\bibitem{v10}
J.~Yang and M.-H. Yang, ``Top-down visual saliency via joint crf and dictionary
  learning,'' \emph{CVPR}, 2012.

\bibitem{v12}
N.~D.~B. Bruce and J.~K. Tsotsos, ``Saliency, attention, and visual search: An
  information theoretic approach,'' \emph{Journal of Vision}, 2009.

\bibitem{v13}
X.~Hou and L.~Zhang, ``Saliency detection: A spectral residual approach,''
  \emph{CVPR}, 2007.

\bibitem{v14}
L.~Zhang, M.~H. Tong, T.~K. Marks, H.~Shan, and G.~W. Cottrell, ``Sun: A
  bayesian framework for saliency using natural statistics,'' \emph{Journal of
  Vision}, 2008.

\bibitem{v15}
H.~J. Seo and P.~Milanfar, ``Static and space-time visual saliency detection by
  self-resemblance,'' \emph{Journal of Vision}, 2009.

\bibitem{v16}
N.~Murray, M.~Vanrell, X.~Otazu, and C.~A. Parraga, ``Saliency estimation using
  a non-parametric low-level vision model,'' \emph{CVPR}, 2011.

\bibitem{v17}
E.~Erdem and A.~Erdem, ``Visual saliency estimation by nonlinearly integrating
  features using region covariances,'' \emph{Journal of Vision}, 2013.

\bibitem{s1}
R.~Achanta, S.~Hemami, F.~Estrada, and S.~Susstrunk, ``Frequency-tuned salient
  region detection,'' \emph{CVPR}, 2009.

\bibitem{s2}
M.-M. Cheng, N.~J. Mitra, X.~Huang, P.~H.~S. Torr, and S.~M. Hu, ``Global
  contrast based salient region detection,'' \emph{IEEE PAMI}, 2015.

\bibitem{s3}
K.~Y. Chang, T.~L. Liu, H.~T. Chen, and S.~H. Lai, ``Fusing generic objectness
  and visual saliency for salient object detection,'' \emph{ICCV}, 2011.

\bibitem{s4}
T.~Liu, Z.~Yuan, J.~Sun, J.~Wang, N.~Zheng, X.~Tang, and H.~Shum, ``Learning to
  detect a salient object,'' \emph{IEEE PAMI}, 2011.

\bibitem{s5}
Y.~Lu, W.~Zhang, H.~Lu, and X.~Y. Xue, ``Salient object detection using
  concavity context,'' \emph{ICCV}, 2011.

\bibitem{s7}
F.~Perazzi, P.~Kr{\"a}henb{\"u}hl, Y.~Pritch, and A.~Hornung, ``Saliency
  filters: Contrast based filtering for salient region detection,''
  \emph{CVPR}, 2012.

\bibitem{s9}
S.~Goferman, L.~Zelnik-Manor, and A.~Tal, ``Context-aware saliency detection,''
  \emph{IEEE PAMI}, 2012.

\bibitem{s10}
X.~Shen and Y.~Wu, ``A unified approach to salient object detection via low
  rank matrix recovery,'' \emph{CVPR}, 2012.

\bibitem{s11}
C.~Yang, L.~Zhang, H.~Lu, X.~Ruan, and M.-H. Yang, ``Saliency detection via
  graph-based manifold ranking,'' \emph{CVPR}, 2013.

\bibitem{s12}
B.~Jiang, L.~Zhang, H.~Lu, C.~Yang, and M.-H. Yang, ``Saliency detection via
  absorbing markov chain,'' \emph{ICCV}, 2013.

\bibitem{s14}
Q.~Yan, L.~Xu, J.~Shi, and J.~Jia, ``Hierarchical saliency detection,''
  \emph{CVPR}, 2013.

\bibitem{s15}
P.~Jiang, H.~Ling, J.~Yu, and J.~Peng, ``Salient region detection by ufo:
  Uniqueness, focusness and objectness,'' \emph{ICCV}, 2013.

\bibitem{s16}
H.~Jiang, J.~Wang, Z.~Yuan, Y.~Wu, N.~Zheng, and S.~Li, ``Salient object
  detection: A discriminative regional feature integration approach,''
  \emph{CVPR}, 2013.

\bibitem{s17}
L.~Mai, Y.~Niu, and F.~Liu, ``Saliency aggregation: A data-driven approach,''
  \emph{CVPR}, 2013.

\bibitem{s18}
S.~Lu, V.~Mahadevan, and N.~Vasconcelos, ``Learning optimal seeds for
  diffusion-based salient object detection,'' \emph{CVPR}, 2014.

\bibitem{s19}
R.~Liu, J.~Cao, Z.~Lin, and S.~Shan, ``Adaptive partial differential equation
  learning for visual saliency detection,'' \emph{CVPR}, 2014.

\bibitem{s20}
J.~Kim, D.~Han, Y.-W. Tai, and J.~Kim, ``Salient region detection via
  high-dimensional color transform,'' \emph{CVPR}, 2014.

\bibitem{s21}
W.~Zhu, S.~Liang, Y.~Wei, and J.~Sun, ``Saliency optimization from robust
  background detection,'' \emph{CVPR}, 2014.

\bibitem{s22}
Z.~Ren, Y.~Hu, L.-T. Chia, and D.~Rajan, ``Improved saliency detection based on
  superpixel clustering and saliency propagation,'' \emph{ACM Multimedia},
  2010.

\bibitem{s23}
M.-M. Cheng, J.~Warrell, W.-Y. Lin, S.~Zheng, V.~Vineet, and N.~Crook,
  ``Efficient salient region detection with soft image abstraction,''
  \emph{ICCV}, 2013.

\bibitem{s24}
J.~Zhang and S.~Sclaroff, ``Saliency detection: A boolean map approach,''
  \emph{ICCV}, 2013.

\bibitem{s28}
X.~Li, H.~Lu, L.~Zhang, X.~Ruan, and M.-H. Yang, ``Saliency detection via dense
  and sparse reconstruction,'' \emph{ICCV}, 2013.

\bibitem{grab}
Q.~Wang, W.~Zheng, and R.~Piramuthu, ``Grab: Visual saliency via novel graph
  model and background priors,'' \emph{CVPR}, 2016.

\bibitem{EQCUTS}
A.~Aytekin, E.~C. Ozan, S.~Kiranyaz, and M.~G. Tampere, ``Visual saliency by
  extended quantum cuts,'' \emph{ICIP}, 2015.

\bibitem{GP}
P.~Jiang, N.~Vasconcelos, and J.~Peng, ``Generic promotion of diffusion-based
  salient object detection,'' \emph{ICCV}, 2015.

\bibitem{SP}
C.~Gong, D.~Tao, W.~Liu, S.~J. Maybank, M.~Fang, K.~Fu, and J.~Yang, ``Saliency
  propagation from simple to difficult,'' \emph{CVPR}, 2015.

\bibitem{RW}
C.~Li, Y.~Yuan, W.~Cai, Y.~Xia, and D.~D. Feng, ``Robust saliency detection via
  regularized random walks ranking,'' \emph{CVPR}, 2015.

\bibitem{a3}
R.~Achanta, A.~Shaji, K.~Smith, A.~Lucchi, P.~Fua, and S.~S{\"u}sstrunk, ``Slic
  superpixels compared to state-of-the-art superpixel methods,'' \emph{IEEE
  PAMI}, 2012.

\bibitem{ERS}
M.~Y. Liu, O.~Tuzel, S.~Ramalingam, and R.~Chellappa, ``Entropy rate superpixel
  segmentation,'' \emph{CVPR}, 2011.

\bibitem{a4}
R.~R. Coifman and S.~Lafon, ``Diffusion maps,'' \emph{Applied and Computational
  Harmonic Analysis}, 2006.

\bibitem{a1}
A.~Ng, M.~Jordan, and Y.~Weiss, ``On spectral clustering: Analysis and an
  algorithm,'' \emph{NIPS}, 2002.

\bibitem{a2}
U.~Luxburg, ``A tutorial on spectral clustering,'' \emph{Statistics and
  Computing}, 2007.

\bibitem{a9}
B.~Nadler and M.~Galun, ``Fundamental limitations of spectral clustering,''
  \emph{NIPS}, 2006.

\bibitem{a8}
J.~A. Tropp and A.~C. Gilbert, ``Signal recovery from random measurements via
  orthogonal matching pursuit,'' \emph{IEEE Transactions on Information
  Theory}, 2007.

\bibitem{s31}
R.~Margolin, A.~Tal, and L.~Zelnik-Manor, ``What makes a patch distinct?''
  \emph{CVPR}, 2013.

\bibitem{a10}
A.~Borji, M.-M. Cheng, H.~Jiang, and J.~Li, ``Salient object detection: A
  benchmark,'' \emph{IEEE TIP}, 2015.

\bibitem{deep1}
P.~Zhang, D.~Wang, H.~Lu, H.~Wang, and X.~Ruan, ``Amulet: Aggregating
  multi-level convolutional features for salient object detection,''
  \emph{ICCV}, 2017.

\bibitem{deep2}
G.~Li and Y.~Yu, ``Deep contrast learning for salient object detection,''
  \emph{CVPR}, 2016.

\end{thebibliography}

%

\begin{IEEEbiography}[{\includegraphics[width=1in,height=1.25in,clip,keepaspectratio]{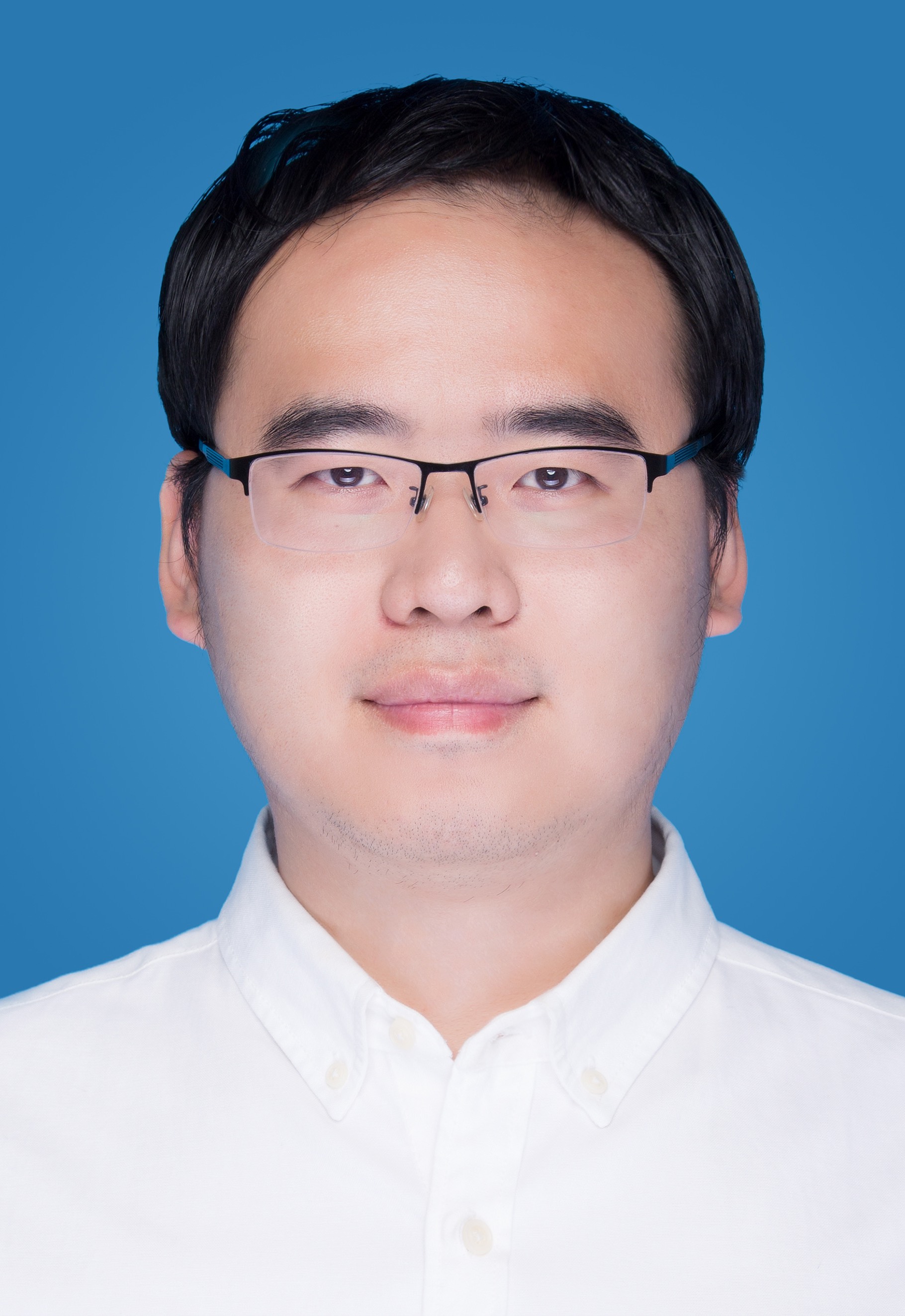}}]{Peng Jiang}
received the BS and PhD degrees in computer science and technology from Shandong University, China,
in 2010 and 2016, respectively. Currently, He is a
Lecturer with the School of Qilu Transportation, Shandong University, China. His research interests include computer vision and machine learning.
\end{IEEEbiography}

\begin{IEEEbiography}[{\includegraphics[width=1in,height=1.25in,clip,keepaspectratio]{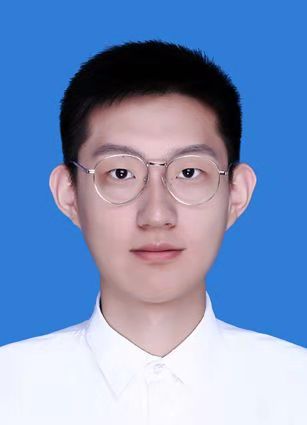}}]{Zhiyi Pan}
received the BS degree from Shandong University, China, in 2018. He is currently pursuing
the MS degree in computer science and technology with
Shandong University, China. His research
interests include computer vision and machine learning.
\end{IEEEbiography}


\begin{IEEEbiography}[{\includegraphics[width=1in,height=1.25in,clip,keepaspectratio]{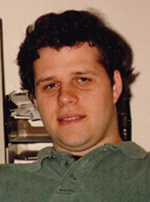}}]{Nuno Vasconcelos}
received his PhD from the Massachusetts Institute of Technology in 2000. From 2000 to 2002, he was a member of the research staff at the Compaq Cambridge Research
Laboratory. In 2003, he joined the Department of Electrical and Computer Engineering at the University of California, San Diego, where he is the head of the Statistical Visual
Computing Laboratory. His work spans various areas, including computer vision, machine learning, signal processing, and multimedia systems.
\end{IEEEbiography}

\begin{IEEEbiography}[{\includegraphics[width=1in,height=1.25in,clip,keepaspectratio]{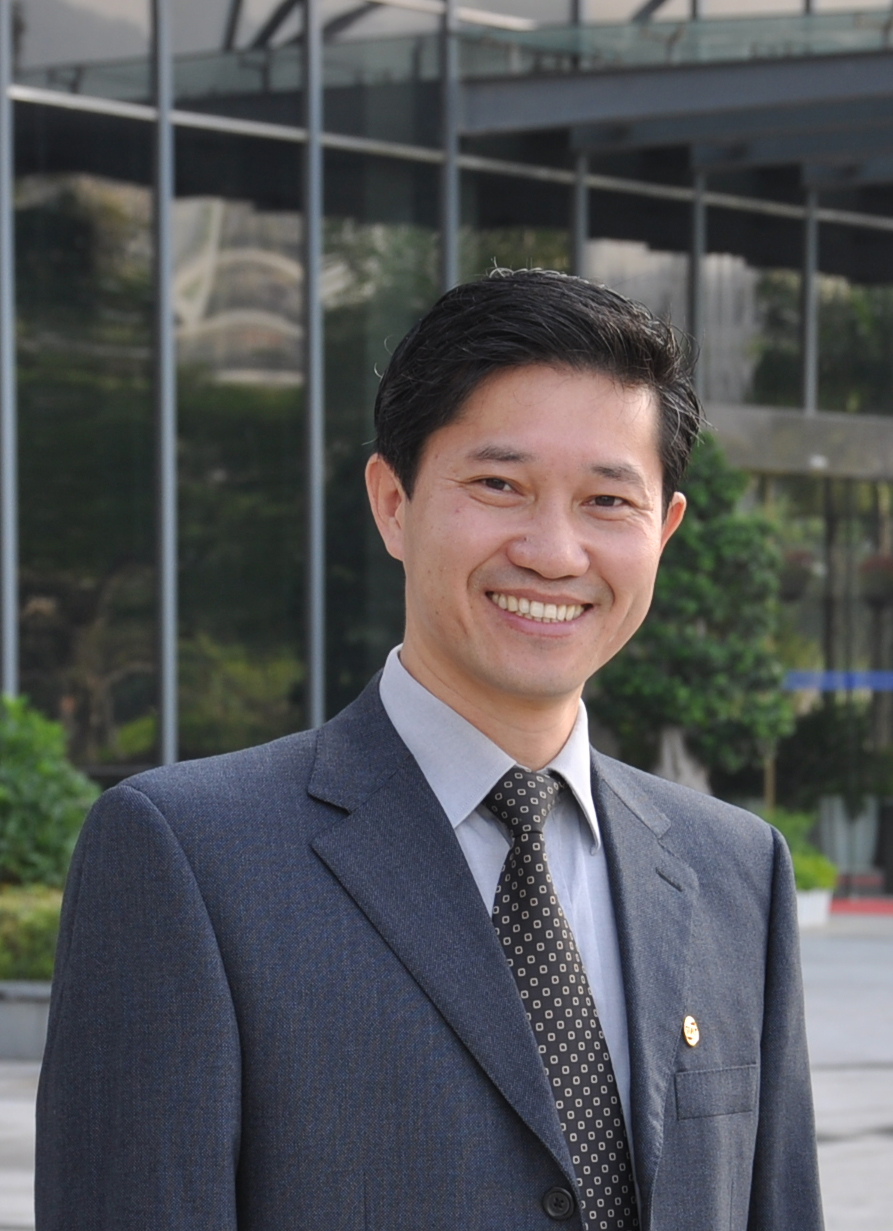}}]{Baoquan Chen}
is a Chair Professor of Peking University, where he is the Executive Director of the Center on Frontiers of Computing Studies. Prior to the current post, he was the Dean of School of Computer Science and Technology  at Shandong University, and the founding director of the Visual Computing Research Center, Shenzhen Institute of Advanced Technology (SIAT), Chinese Academy of Sciences (2008-2013), and a faculty member at Computer Science and Engineering at the University of Minnesota at Twin Cities (2000-2008). His research interests generally lie in computer graphics, visualization, and human-computer interaction, focusing specifically on large-scale city modeling, simulation and visualization. He has published more than 100 papers in international journals and conferences, including two dozens or so papers in SIGGRAPH and SIGGRAPH Asia. Chen received an MS in Electronic Engineering from Tsinghua University, Beijing (1994), and a second MS (1997) and then PhD (1999) in Computer Science from the State University of New York at Stony Brook. Chen is the recipient of the Microsoft Innovation Excellence Program 2002, the NSF CAREER award 2003, McKnight Land-Grant Professorship for 2004, IEEE Visualization Best Paper Award 2005, and NSFC "Outstanding Young Researcher" program in 2010. Chen served as conference co-chair of IEEE Visualization 2005, and served as the conference chair of SIGGRAPH Asia 2014.
\end{IEEEbiography}

\begin{IEEEbiography}[{\includegraphics[width=1in,height=1.25in,clip,keepaspectratio]{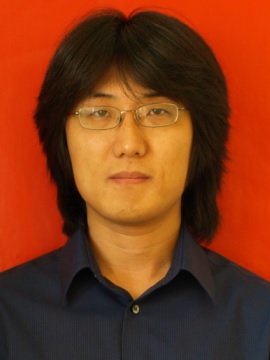}}]{Jingliang Peng}
received the PhD degree in electrical engineering from the University of Southern California in 2006, the BS and MS degrees in computer science from Peking University in 1997 and 2000, respectively. Currently, he is a professor at the School of Software, Shandong University, China. His research interest mainly resides in digital geometry processing and digital image/video analysis.
\end{IEEEbiography}



\end{document}